\documentclass{article}

    \PassOptionsToPackage{numbers, compress}{natbib}

\usepackage{wrapfig, makecell}

    \usepackage[final]{neurips_2022}

\usepackage[utf8]{inputenc} 
\usepackage[T1]{fontenc}    
\usepackage{hyperref}       
\usepackage{url}            
\usepackage{booktabs}       
\usepackage{amsfonts}       
\usepackage{nicefrac}       
\usepackage{microtype}      
\usepackage{xcolor}         
\usepackage{graphicx}
\usepackage{floatrow}
\usepackage{enumitem}
\usepackage{wrapfig}
\usepackage{amsmath}

\title{AutoLink: Self-supervised Learning of Human Skeletons and Object Outlines by Linking Keypoints}

\author{Xingzhe He \hspace{5mm} Bastian Wandt \hspace{5mm} Helge Rhodin \\
University of British Columbia\\
{\tt\small \{xingzhe, wandt, rhodin\}@cs.ubc.ca}}

\usepackage{soul} 

\newif\ifdraft
\drafttrue

\newcommand{\R}{\mathbb{R}}

\newcommand{\comment}[1]{}

\newcommand{\parag}[1]{{\bf{#1}}}

\newcommand{\vd}{\mathbf{d}}

\newcommand{\vk}{\mathbf{k}}

\newcommand{\vp}{\mathbf{p}}

\newcommand{\vw}{\mathbf{w}}
\newcommand{\vx}{\mathbf{x}}

\newcommand{\vz}{\mathbf{z}}

\newcommand{\mF}{\mathbf{F}}

\newcommand{\mH}{\mathbf{H}}
\newcommand{\mI}{\mathbf{I}}

\newcommand{\mS}{\mathbf{S}}

\newcommand{\cD}{\mathcal D}

\newcommand{\cG}{\mathcal G}

\newcommand{\cL}{\mathcal L}

\newcommand{\cN}{\mathcal N}

\begin{document}

\maketitle
\begin{abstract}
Structured representations such as keypoints are widely used in pose transfer, conditional image generation, animation, and 3D reconstruction.
However, their supervised learning requires expensive annotation for each target domain.
We propose a self-supervised method that learns to disentangle object structure from the appearance with a graph of 2D keypoints linked by straight edges. 
Both the keypoint location and their pairwise edge weights are learned, given only a collection of images depicting the same object class. 
The resulting graph is interpretable, for example, AutoLink recovers the human skeleton topology when applied to images showing people.
Our key ingredients are i) an encoder that predicts keypoint locations in an input image, ii) a shared graph as a latent variable that links the same pairs of keypoints in every image, iii) an intermediate edge map that combines the latent graph edge weights and keypoint locations in a soft, differentiable manner, and iv)
an inpainting objective on randomly masked images.
Although simpler, 
AutoLink outperforms existing self-supervised methods on the established keypoint and pose estimation benchmarks 
and paves the way for structure-conditioned generative models on more diverse datasets.
Project website: \url{https://xingzhehe.github.io/autolink/}.
\end{abstract}
\section{Introduction}

Object structure representations are widely used in modern computer graphics and computer vision techniques, including keypoints for image generation \cite{ma2017pose, ma2018disentangled, siarohin2018deformable} 
and skeletons for 3D reconstruction \cite{feng2018joint, jackson2017large, yu2017bodyfusion, pavlakos2019expressive, su2021anerf}.
However, the structure is usually supervised on large annotated datasets~\cite{lin2014microsoft, andriluka14cvpr, PoseTrack, mono3dhp2017} or via hand-crafted parametric models~\cite{SMPL2015, SMPLX2019, Xu2020ghum, osman2020star, blanz1999morphable, FLAME:SiggraphAsia2017}. Neither approach generalizes well to new domains and both require additional manual annotation whenever more detail is needed~\cite{Guler2018DensePose}. 

Our goal is to reconstruct the keypoint locations of an object by learning from an unlabelled image collection, thereby sidestepping the generalization problem.
Our key idea is to leverage that the same object shares the same topology by introducing an explicit graph that links the same pairs of keypoints in all instances.
By contrast, existing self-supervised keypoint learning methods model objects as a set of independent parts.
Their consistency over different instances of the same object is encouraged by either enforcing parts to follow hand-crafted image transformations~\cite{thewlis2017unsupervised, zhang2018unsupervised, jakab2018unsupervised, lorenz2019unsupervised, hung2019scops, liu2021unsupervised} or by adding implicit bias in the network architecture that encodes such spatial equivariances~\cite{he2022ganseg, he2021latentkeypointgan}.
Only~\cite{jakab2020self, schmidtke2021unsupervised} use an explicit skeleton representation, but both require predefined topology, rely on video input, and \cite{jakab2020self} is trained in a CycleGAN setting that still requires manually labeled examples.

We propose a simple yet effective method to learn both the keypoints and their links without supervision in terms of a sparse graph serving two purposes. 
First, the graph acts as a bottleneck that can only store structural information disentangled from appearance. 
Second, it forms a constraint that associates observations across training images. We enforce the same topology across instances of the same class by learning a shared graph with a single set of edge weights.
In the absence of labels, we train AutoLink with only an autoencoder reconstruction objective. Since the graph bottleneck should not model appearance, we additionally feed the decoder with the input image after masking the majority of its pixels.
In turn, it is important for the disentanglement that the heavily masked image contains appearance without leaking structural information. This is the case as inpainting methods are unable to infer the original image precisely from only sparse pixel colors \cite{zheng2019pluralistic, yu2021diverse}
unless conditioning on structure representations, such as edge maps~\cite{xiong2019foreground, nazeri2019edgeconnect, li2019progressive, lahiri2020prior, han2019finet}. 
Therefore, by forcing the autoencoder to reconstruct the original image, the detector converges to generate representative structures of images.

\begin{figure}[t]
\begin{center}
  \includegraphics[width=0.98\textwidth]{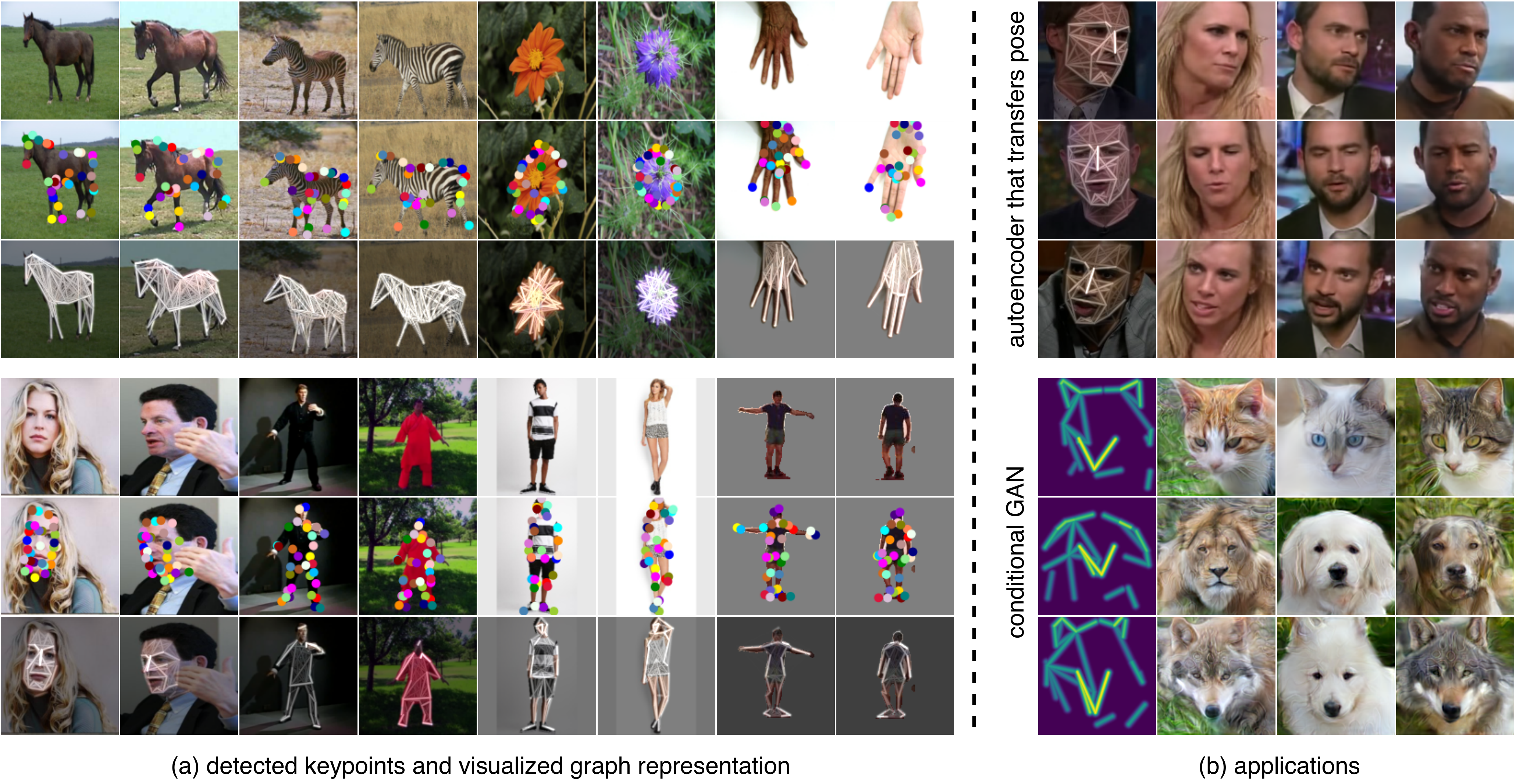}
\end{center}
    \caption{\textbf{Teaser.}
    (a) AutoLink applies to diverse collections of images and automatically yields keypoints linked to a graph without ground truth. It recovers animal and human poses and object shapes in settings where previous methods struggle, including cluttered backgrounds, structured stripe textures, articulated fingers, and detailed faces. 
    (b) Example applications are conditional image generation with autoencoders (top) and GANs (bottom) that are driven by the learned keypoints.}
\label{fig:teaser}
\end{figure}

We demonstrate on 4 benchmarks that the trained detector has a significantly improved keypoint localization accuracy and on 6 additional datasets that it applies to a broader set of images spanning portraits, persons, animals, hands, and flowers, which we attribute to the explicit modeling of links in the graph. 
Figure~\ref{fig:teaser} shows the diverse set of image domains it applies to, including challenging textures and uncontrolled background, 
how both skeleton representations as well as object outlines are learned by varying the number of keypoints,
and exemplifies applications to controlled image generation. 

\textbf{Ethics - Risks.} \label{sec:ethics} The estimated keypoints and edges could be abused for deep fakes as the driving signal for generative models or for unwanted surveillance applications. However, our method works towards improved generality, including objects and animals, and does not improve upon supervised models that already exist in high detail for humans.
\textbf{Benefits.} Since our method is entirely self-supervised, it can be applied to a diverse set of persons, objects, animals, or situations that have not yet been labeled.

\section{Related Work}

Most representation learning methods focus on generic feature vectors for entire images to initialize deep networks for improved object classification \cite{wu2018unsupervised, chen2020simple, oord2018representation, he2020momentum, he2021masked}.
By contrast, our method introduces explicit object structure. We review the most related approaches in the following.

\noindent\parag{Self-supervised Keypoints Detection.} The most common idea to discover keypoints in an unsupervised manner is to rely on the notion that keypoints move as the image changes.
Various constraints have been used
to enforce that the keypoints follow a known transformation,
including
view changes in a multi-view recording \cite{suwajanakorn2018discovery, rhodin2018unsupervised, Rhodin_2019_CVPR} or the natural motion in videos \cite{dundar2020unsupervised, siarohin2019animating, kulkarni2019unsupervised, minderer2019unsupervised, dong2018supervision, kim2019unsupervised, jakab2020self}.
When only single images are available, artificial image deformation is applied, either from randomized
\cite{thewlis2017unsupervised, zhang2018unsupervised, jakab2018unsupervised, lorenz2019unsupervised} or learned \cite{wu2019transgaga, xu2020unsupervised} transformations within a pre-defined deformation space. 
However, 
learned keypoints may model the background \cite{zhang2018unsupervised, siarohin2019animating} and struggle with large pose variation \cite{hung2019scops} as image deformations do not separate foreground from background, which are usually tuned for each dataset
and bound to be small.
Most models leverage multi-branch network architectures to encode the structure and appearance separately and utilize multiple losses that need to be balanced. By contrast, we do not apply artificial transformations, use a single branch, and a single loss which eases and stabilizes training.
To overcome the need for artificial image deformation, He et al. \cite{he2021latentkeypointgan, he2022ganseg} exploit GANs to generate images along with corresponding keypoints and later use them to train a detector. However, this leads to even more complex network architectures and comes with instabilities in GAN training, limiting their applicability to complex objects like human bodies. \looseness=-1

\noindent\parag{Skeleton Representations.} Bone maps representing the keypoints connectivity as affinity fields \cite{cao2017realtime} or via explicit offsets \cite{papandreou2018personlab} are used in supervised 
human, animal, and object pose estimation.
We use a similar edge map representation but learn both the location and linking from scratch without annotations.
In the weakly-supervised setting, Jakab et al. \cite{jakab2020self} exploit CycleGAN \cite{zhu2017unpaired} to translate between image and edge maps. The graph connectivity is predefined to the human skeleton and edges are supervised by a large dataset of unpaired ground truth object edges, which can come from a different dataset but are manually annotated. Schmidtke et al. \cite{schmidtke2021unsupervised} overcome the manual labeling by deforming a template skeleton. However, they both require the known connectivity of the keypoints and videos for training while ours learns both the keypoints and connectivity from a collection of single images. Noguchi et al. \cite{noguchi2021watch} generate a skeleton heuristically by linking the centers of part-wise learned Signed Distance Fields \cite{malladi1995shape}. However, they require videos without the background of the same object, and the learned skeleton does not generalize to other objects of the same class.

\noindent\parag{Object Sketch Learning.} Sketches are made of strokes drawn by a pen. It is a concise and abstract representation, which can be used in object recognition \cite{Yu2015SketchaNetTB, xu2021multigraph} and image retrieval \cite{wang2015sketch, qian2016sketch, xu2018sketchmate}. There are two common sketch representations used in neural models~\cite{Xu2022DeepLF}:
black-white raster images \cite{wang2019learning, SasakiCGI2018learning,  YiLLR19APDrawingGAN}, often used for image-to-image translation \cite{isola2017image, zhu2017unpaired, pang2018deep, kampelmuhler2020synthesizing}
and sequences of points (pen coordinates) \cite{ha2018a, eitz2012hdhso, sangkloy2016the}, which is usually used by recurrent generation models \cite{ha2018a, chen2017sketch, Cao2019ai, kaiyrbekov2019deep, das2020beziersketch, ge2021creative}. This graph representation is similar to ours. However, instead of learning to mimic human drawings, ours directly predicts both the keypoints and their connectivity on real natural images.

\noindent\parag{Structure-enhanced Image Inpainting.} When key parts are missing in an image, e.g., eyes on faces or arms of humans, it is hard for inpainting networks to imagine the content accurately from scratch.
Therefore, additional structural cues are detected to guide the subsequent image generation. 
The cues can be supervised segmentation masks \cite{zhao2021prior, liao2020guidance, han2019finet, song2018spg}, foreground contours \cite{xiong2019foreground}, and landmarks \cite{lahiri2020prior, zhang2021face, yang2020generative}, or automatically extracted
 edges \cite{nazeri2019edgeconnect, li2019progressive, jie2020inpainting, xu2020anedge, cao2021learning} and low-frequency image components \cite{wang2020image, ren2019structureflow}. 
Our reconstruction objective can be seen as such two-stage inpainting, but self-supervised and with the image edges replaced with the learned graph edge representation. \looseness=-1

\noindent\parag{Self-supervised Foreground Segmentation.} Traditional methods use color \cite{zhu2014saliency}, contrast \cite{cheng2011global}, and hand-crafted features \cite{jiang2013salient} to cluster foreground pixels. 
A recent trend is exploiting inpainting techniques to segment the foreground. Chen et al.~\cite{chen2019unsupervised} and Arandjelovi{\'c} et al.~\cite{arandjelovic2019object} use a GAN to inpaint the background at the predicted segmentation mask, assuming that the object texture can be changed without changing the data distribution due to the independence of foreground and background. Yang et al.~\cite{yang2019unsupervised} propose Contextual Information Separation (CIS),
a general objective to segment the foreground by maximizing the error of inpainting the mask and its complement. It was first applied to optical flow maps and subsequently to RGB images by \cite{savarese2021information,yang2021dystab, katircioglu2021self}. 
When the object is small compared to the background, an additional object detection module 
\cite{katircioglu2021self,crawford2019spatially}
or  multi-view \cite{katircioglu2021human} information is required.
Different from these previous methods, we utilize a form of inpainting to learn sparse keypoints instead of segmentation. Our learned edges form a sparse foreground shape, but further extensions would be necessary to transfer from the edge maps to the boundary-aligned foreground segmentation.

\section{Method} \label{sec:method}

\begin{figure}[t]
\begin{center}
  \includegraphics[width=1.\textwidth]{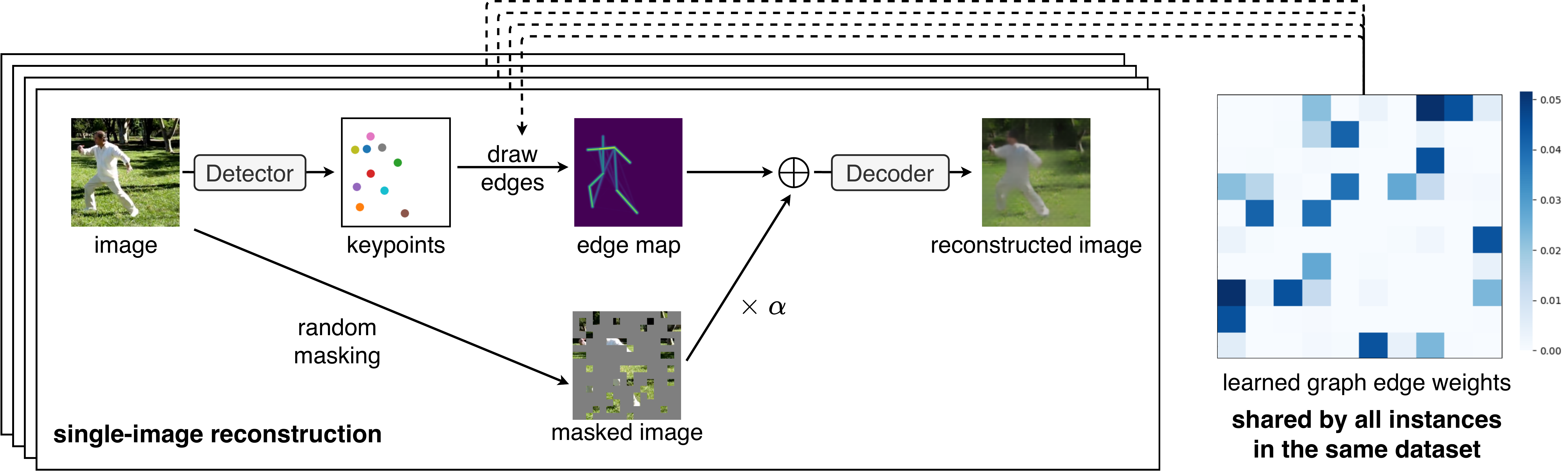}
\end{center}
   \caption{\textbf{Overview.} Given an image, we detect keypoints and draw differentiable edges between keypoints according to the learned graph edge weights that is visualized as a color matrix. The method is self-supervised in that the latent edge map and keypoints are learned by reconstructing the masked input images. Note that keypoints are image specific and edge maps are shared.}
\label{fig:overview}
\end{figure}

We leverage that the objects in the dataset share the same topology and can be represented as a graph that connects keypoints by a shared set of edges. To learn the keypoints and edges, we design an autoencoder that aims to accurately reconstruct the input image, with the graph as the intermediate representation. To encode the input image into a graph, we detect the keypoints and create an edge heatmap based on learnable edge weights. To mostly obtain appearance information, we mask out the majority of the image, which randomizes the structure information and reduces the remainder to a very low  level.
The edge heatmap is combined with the masked image to reconstruct the original image.
Since the missing structure is important to reconstructing the original image, the network is forced to learn the structure of the object in a self-supervised manner. 
Figure~\ref{fig:overview} shows an overview of our method. 

Formally, given an image $\mI\in\R^{H\times W\times 3}$ with height $H$ and width $W$ we aim to learn a set of keypoints $\{\vk_i\}_{i=1}^K$, where $\vk_i\in[-1,1]\times[-1,1]\subset \R^2$ is the normalized keypoint coordinate, and $K$ is the number of keypoints. 
We use a ResNet with upsampling \cite{xiao2018simple} to detect keypoints. Afterward, we draw a differentiable edge \cite{mihai2021differentiable} between each pair of keypoints (details below). 
The edge map $\mS\in\R^{H\times W}$ is concatenated along the channel dimension with the randomly masked image $\mI_m\in\R^{H\times W\times 3}$ and fed into a UNet \cite{ronneberger2015u} to obtain the reconstructed image $\mI'$. The detailed network architectures can be found in Appendix~\ref{supp:archi}. \looseness=-1

\subsection{Image Structure Representation}

In this section, we introduce the generation of keypoints and the edge map from the image.
Let $\mH\in\R^{H\times W\times K}$ be the $K$ heatmaps generated by a ResNet with upsampling \cite{xiao2018simple} from the image $\mI$. The keypoint $\vk_i$ is calculated by the differentiable soft-argmax function,
\begin{equation}
    \vk_i = \sum_\vp \frac{ \exp(\mH(\vp)) }{\sum_\vp (\exp\mH(\vp))} \vp,
\end{equation}
where $\vp\in [-1,1]\times[-1, 1]$ is the normalized pixel coordinates.

Given two keypoints $\vk_i, \vk_j$, we draw a differentiable edge map $\mS_{ij}$, where values are 1 on the edge linked by the two keypoints and decrease exponentially based on the distance to the line. Formally, the edge map $\mS_{ij}$ is a Gaussian extended along the line \cite{mihai2021differentiable}, defined as
\begin{equation}
    \mS_{ij}(\vp) = \exp\left(d^2_{ij}(\vp) / \sigma^2\right), 
\end{equation}
where $\sigma$ is a hyperparameter controlling the thickness of the edge, and $d_{ij}(\vp)$ is the $L_2$ distance between the pixel $\vp$ and the edge drawn by keypoints $\vk_i$ and $\vk_j$,
\begin{equation}
\vd_{ij}(\vp) = \left\{
\begin{aligned}
& \|\vp-\vk_i\|_2 &\text{ if } t\leq 0, \\
& \|\vp-((1-t)\vk_i + t\vk_j)\|_2 &\text{ if } 0<t<1, \\
& \|\vp-\vk_j\|_2 &\text{ if } t\geq 1,
\end{aligned}
\right.
\quad \text{where}\quad
t = \frac{(\vp-\vk_i)\cdot (\vk_j-\vk_i)}{\|\vk_i-\vk_j\|^2_2}.
\end{equation}

We assign a weight $w_{ij}>0$ to each edge, which is enforced to be positive by SoftPlus \cite{dugas2000incorporating}. This weight is learned during training and shared across all object instances in a dataset. 
Finally, we take the maximum at each pixel of the heatmaps to obtain the final edge map $\mS\in\R^{H\times W}$,
\begin{equation}
   \mS(\vp)  = \max_{ij}w_{ij}\mS_{ij}(\vp).
   \label{eq:max_heatmap}
\end{equation}
Taking the maximum at each pixel avoids the entanglement of the edge weights and the convolution kernel weights, which is further explained in Section~\ref{sec:ablation}.

\subsection{Image Reconstruction}
The masked image $\mI_m$ is generated by first uniformly dividing the image $\mI$ into a $16\times 16$ grid, and randomly masking out 80\% of the grid cells, similar to \cite{he2021masked}. 
We concatenate the masked image with the edge map and feed it into a UNet decoder \cite{ronneberger2015u} to reconstruct the original image,
\begin{equation}
    \mI'=\text{Decoder}(\alpha\mI_m \oplus \mS)
    \label{eq:recon}
\end{equation}
where $\oplus$ means concatenation along the channel dimension and $\alpha$ is a learnable parameter that compensates for the change of the edge weight magnitude during training. $\alpha$ is initialized to 1. We found this parameter to be helpful in training stability.
Different to \cite{he2021masked}, we condition on an edge map.
Different to \cite{jakab2020self}, we have no ground truth for the edge map. Our edge map is an unobserved latent variable. Thus we only minimize the difference of the original image and the reconstructed image by the perceptual loss~\cite{johnson2016perceptual},
\begin{equation}
    \cL=\frac{1}{N}\sum_{i=1}^N\|\Gamma(I_i)-\Gamma(I'_i)\|^2_2
\end{equation}
where $N$ is the number of examples and $\Gamma$ is the feature extractor.
The perceptual loss is believed to measure the structure similarity \cite{johnson2016perceptual, gatys2016image, dosovitskiy2016generating}, and leads to more robust training \cite{jakab2018unsupervised, jakab2020self}.

\subsection{Implementation Details} \label{sec:implementation_details}
We use the Adam optimizer \cite{KingmaB14} with a learning rate of $10^{-4}$ with $\beta_1=0.9$, $\beta_2=0.99$. The batch size is 64. We train for 20k iterations. It takes 3 hours to train on a single V100 GPU. All images are resized to $128\times 128$. The learning rate for the edge weights is multiplied by 512 due to the small gradient of SoftPlus \cite{dugas2000incorporating} when the value is close to 0.
To show the robustness of our model, we report all experiments on the sampling strategy of masking 80\% of the $16\times 16$ patches. 
We perform experiments with the same edge thickness of $\sigma^2=5e-5$ for all benchmark datasets. 
We train 10 times and report the mean and the standard deviation of the evaluation metrics. 
Although it already outperforms other work in most experiments, we also tune thicknesses to each individual dataset, as others did for their hyperparameters, which further improves the results. The tuned thicknesses can be found in Appendix~\ref{supp:ablation_test}.
The only other hyperparameter is the number of keypoints, which we set to that of the established benchmarks for quantitative comparisons, ranging from 4 to 32 points.
\section{Experiments} \label{experiment}

\begin{figure}[t]
\centering
  \resizebox{0.98\linewidth}{!}{%
\begin{tabular}{cc}
\rotatebox{90}{$\phantom{xx}$ LatentKeypointGAN \cite{he2021latentkeypointgan}}&
\includegraphics[]{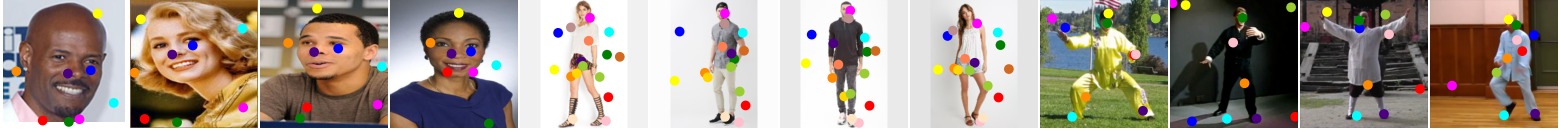}\\
\rotatebox{90}{$\phantom{x}$ \huge{GANSeg} \cite{he2022ganseg}}&
\includegraphics[]{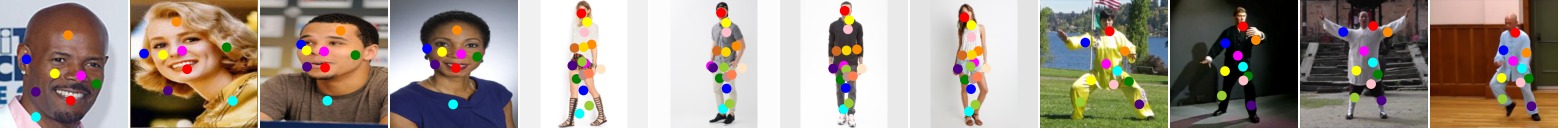}\\
\rotatebox{90}{$\phantom{xxxxxxxx}$ \huge Ours} &
\includegraphics[]{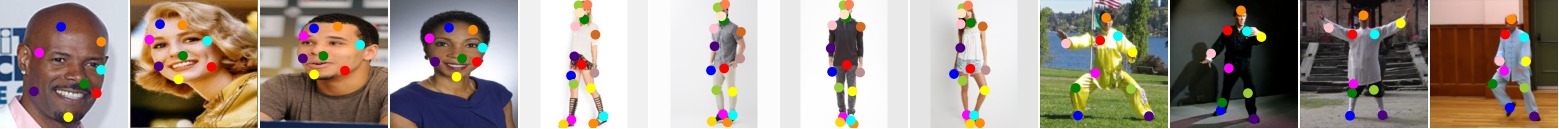}\\
\rotatebox{90}{$\phantom{xx}$ \huge  Ours (edges)} &
\includegraphics[]{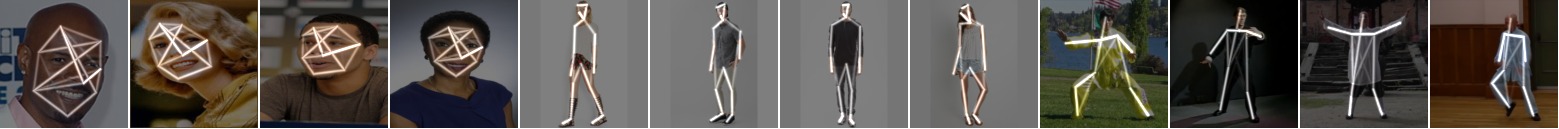}
\end{tabular}}%
\caption{\textbf{Qualitative comparison on detected keypoints}. Our model is more robust on wild face poses, and depicts more details on human bodies compared to \cite{he2021latentkeypointgan} and \cite{he2022ganseg}. For example, the feet poses in the middle four images are clearly detected.}
\label{fig:qualitative}
\end{figure}

In this section, we compare our results to the related methods, showing that our model is simple yet effective. Besides, we perform a number of ablation studies on hyperparameters and algorithm variants, exhibiting the robustness of our model and justifying the necessity of every model component.

\begin{table}[t]
\centering
\caption{\textbf{Landmark detection on CelebA}. The metric is the landmark regression (without bias) error in terms of $L_2$ distance normalized by inter-ocular distance (lower is better). While all methods perform well on aligned CelebA, ours is more robust on Wild CelebA. The sign $\star$ means being reported by \cite{hung2019scops} and $\dagger$ means being reported by \cite{liu2021unsupervised}.} 
\resizebox{0.9\linewidth}{!}{%
\begin{tabular}{llccc} \Xhline{1.5pt}
Method & Type & Aligned (K=10) & Wild (K=4) & Wild (K=8)  \\ \Xhline{1.5pt}

DFF \cite{collins2018deep} by \cite{hung2019scops} & Part Segmentation & - & - & \phantom{$\star$} 31.30\% $\star$ \\
SCOPS \cite{hung2019scops} (w/o saliency)  & Part Segmentation & - & 46.62\% & 22.11\% \\
SCOPS \cite{hung2019scops} (w/ saliency)  & Part Segmentation & - & 21.76\% & 15.01\% \\
Liu et al. \cite{liu2021unsupervised}  & Part Segmentation & - & 15.39\% & 12.26\% \\
Huang et al. \cite{huang2020interpretable} (w/ detailed label)  & Part Segmentation & - & - & \phantom{x}  8.40\% \\
GANSeg \cite{he2022ganseg} & Part Segmentation & 3.98\% & \textbf{12.26}\% & \phantom{x}  \textbf{6.18}\% \\ \hline

Thewlis et al. \cite{thewlis2017unsupervised} & Landmark & 7.95\% & - & \phantom{$\star$} 31.30\% $\star$ \\ 
Zhang et al. \cite{zhang2018unsupervised} & Landmark & 3.46\% & - & \phantom{$\star$} 40.82\% $\star$\\ 
LatentKeypointGAN \cite{he2021latentkeypointgan} & Landmark & 5.85\% & 25.81\% & 21.90\% \\
Lorenz et al. \cite{lorenz2019unsupervised} & Landmark & 3.24\% & \phantom{$\star$} 15.49\% $\dagger$ & \phantom{$\star$} 11.41\% $\dagger$\\
IMM \cite{jakab2018unsupervised} & Landmark &  \textbf{3.19}\% & \phantom{$\star$} 19.42\% $\dagger$ & \phantom{$\star$} \phantom{x} 8.74\% $\dagger$\\
LatentKeypointGAN-tuned \cite{he2021latentkeypointgan} & Landmark & 3.31\% & 12.10\% & \phantom{x} 5.63\% \\
Ours (general) & Landmark &  3.92$\pm$0.69\% & 7.72$\pm$0.47\% & 5.66$\pm$0.29\% \\
Ours (thickness-tuned) & Landmark &  3.54\% & \phantom{x}  \textbf{6.11}\% & \phantom{x}  \textbf{5.24}\% \\ \Xhline{1.5pt}
\end{tabular}}
\label{tab:celeba}
\end{table}

\begin{table}[t]
\centering
\caption{\textbf{Landmark detection on Human Body}. Our model outperforms all the other unsupervised baselines. The metric for each dataset follows the corresponding description in the text. The sign $\dagger$ means being reported by \cite{siarohin2021motion} and the sign $\star$ means being reported by \cite{sandro2020unsupervised}. The number of keypoints is $K=16$ for Human3.6m and DeepFashion and $K=10$ for Taichi.}
\resizebox{0.9\linewidth}{!}{%
\begin{tabular}{llccc} \Xhline{1.5pt}
Method & Supervision & Human3.6m  $\downarrow$ & DeepFashion $\uparrow$ & Taichi $\downarrow$ \\ \Xhline{1.5pt}
Jakab et al. \cite{jakab2020self} & video \& unpaired ground truth & 2.73  & - & - \\ 
Newell et al. \cite{jakab2020self} & paired ground truth & \textbf{2.16}  & - & - \\ \hline
DFF \cite{collins2018deep}& testing dataset & - & - & \phantom{$\star$} 494.48 $\dagger$ \\
SCOPS \cite{hung2019scops}& saliency maps & - & - &  \phantom{$\star$} 411.38 $\dagger$ \\
Siarohin et al. \cite{siarohin2021motion} & videos & - & - & 389.78 \\  
Zhang et al. \cite{zhang2022self} & videos & - & - & \textbf{343.67} \\
Zhang et al. \cite{zhang2018unsupervised} & videos & 4.14  & - & - \\
Schmidtke et al. \cite{schmidtke2021unsupervised} & video \& T-pose template & 3.31 & - & - \\ 
Sun et al. \cite{sun2022self} & videos & \textbf{2.53}$\pm$0.06  & - & - \\ \hline
Thewlis et al. \cite{thewlis2017unsupervised}& unsupervised  & 7.51 & - & - \\
Zhang et al. \cite{zhang2018unsupervised} & unsupervised & 4.91  & - & - \\
LatentKeypointGAN \cite{he2021latentkeypointgan} & unsupervised & -  & 49\% & 437.69\\
Lorenz et al. \cite{lorenz2019unsupervised} & unsupervised & 2.79  & \phantom{$\star$} 57\% $\star$ & - \\ 
GANSeg \cite{he2022ganseg} & unsupervised & - & 59\% & 417.17 \\ 
Ours (general) & unsupervised & 2.81$\pm$0.07 &65$\pm$1.2\% & 337.50$\pm$25.08\\
Ours (thickness-tuned) & unsupervised & \textbf{2.76} & \textbf{66}\% & \textbf{316.10} \\ \Xhline{1.5pt}
\end{tabular}
}
\label{tab:human_body}
\end{table}


\begin{table}[t]
\centering
\caption{\textbf{Landmark detection on CUB Birds}. Our model outperforms most other baselines and achieves comparable results with the ones using ground truth segmentation masks. The metric is the landmark regression (without bias) error of $L_2$ distance normalized by the image size (lower is better). A star $\star$ means being reported by \cite{choudhury2021unsupervised}, $\dagger$ means being reported by \cite{hung2019scops}, and $\ddagger$ means tested by us with their official code; all other numbers are taken from the respective papers. The number of keypoints is $K=10$ for CUB-aligned and $K=4$ for CUB-001, CUB-002, CUB-003, and CUB-all.}
\resizebox{0.9\linewidth}{!}{%
\begin{tabular}{llcccccc} \Xhline{1.5pt}
Method & Supervision & CUB-aligned $\downarrow$ & CUB-001 $\downarrow$ & CUB-002 $\downarrow$ & CUB-003 $\downarrow$ & CUB-all $\downarrow$ \\ \Xhline{1.5pt}
SCOPS \cite{hung2019scops}& GT silhouette & - & \phantom{$\star$}18.3 $\star$ & \phantom{$\star$}17.7 $\star$ & \phantom{$\star$}17.0 $\star$ & \phantom{$\star$} 12.6 $\star$ \\
Choudhury et al. \cite{choudhury2021unsupervised} & GT silhouette & - & \textbf{11.3} & \textbf{15.0} & \textbf{10.6} & \textbf{9.2} \\ \hline
DFF \cite{collins2018deep}& testing dataset & - & \phantom{$\dagger$}22.4$\dagger$ & \phantom{$\dagger$}21.6$\dagger$ & \phantom{$\dagger$}22.0$\dagger$ & - \\
SCOPS \cite{hung2019scops}& saliency maps & - & \textbf{18.5} & \textbf{18.8} & \textbf{21.1} & - \\ \hline
Lorenz et al. \cite{lorenz2019unsupervised} & unsupervised & 3.91  &-&-& -& - \\ 
ULD \cite{zhang2018unsupervised, thewlis2017unsupervised} & unsupervised & - & \phantom{$\dagger$}30.1$\dagger$ & \phantom{$\dagger$}29.4$\dagger$ & \phantom{$\dagger$}28.2$\dagger$ & - \\
Zhang et al. \cite{zhang2018unsupervised} & unsupervised & 5.36 & \phantom{$\ddagger$}26.9$\ddagger$ & \phantom{$\ddagger$}27.6$\ddagger$ & \phantom{$\ddagger$}27.1$\ddagger$ & \phantom{$\ddagger$}22.4$\ddagger$ \\
LatentKeypointGAN \cite{he2021latentkeypointgan} & unsupervised & \phantom{$\ddagger$}5.21$\ddagger$ & \phantom{$\ddagger$}22.6$\ddagger$ & \phantom{$\ddagger$}29.1$\ddagger$ & \phantom{$\ddagger$}21.2$\ddagger$ & \phantom{$\ddagger$}14.7$\ddagger$ \\
GANSeg \cite{he2022ganseg} & unsupervised & \textbf{3.23} & \phantom{$\ddagger$}22.1$\ddagger$ & \phantom{$\ddagger$}22.3$\ddagger$ & \phantom{$\ddagger$}21.5$\ddagger$ & \phantom{$\ddagger$}12.1$\ddagger$\\ 
Ours (general) & unsupervised & 4.15 $\pm$ 0.24 & 20.6 $\pm$ 0.54 & 20.3 $\pm$ 0.96 & 19.7 $\pm$ 0.91 & 11.6 $\pm$ 0.33 \\
Ours (thickness-tuned) & unsupervised & 3.51 & \textbf{20.2} & \textbf{19.2} & \textbf{18.5} &\textbf{11.3} \\ \Xhline{1.5pt}
\end{tabular}
}
\label{tab:cub}
\end{table}

\subsection{Datasets and Evaluation Metrics} \label{sec:datasets}
\noindent\parag{CelebA-aligned} \cite{liu2015faceattributes} contains 200k celebrity faces aligned in center. We follow \cite{thewlis2017unsupervised} splitting it into three subsets: CelebA training set without MAFL (160k images), MAFL training set (19k), MAFL test set (1k). We train our network on the CelebA training set without MAFL. To quantitatively evaluate the consistency of our predicted keypoints, we follow \cite{thewlis2017unsupervised} training a linear regression without bias from our detected keypoints to the ground truth keypoints on the MAFL training set and reporting the mean $L_2$ error normalized by inter-ocular distance on the MAFL test set.

\noindent\parag{CelebA-in-the-wild} \cite{liu2015faceattributes} contains celebrity faces in unconstrained conditions. We follow \cite{hung2019scops} and first split it into three subsets as for CelebA-aligned, and then remove the images where a face covers less than 30\% of the area, which results in 45,609 images for model training, 5,379 with keypoint labels for regression, and 283 for testing. The evaluation metric is the same as CelebA-aligned.

\noindent\parag{Human3.6m} \cite{Ionescu2014human36m} contains human activity videos in static backgrounds. We follow \cite{zhang2018unsupervised} considering six activities (direction, discussion, posing, waiting, greeting, walking), and using subjects 1, 5, 6, 7, 8, 9 for training and 11 for testing. This results in 796,648 images for training and 87,975 images for testing. The evaluation metric is the regressed (without bias) mean $L_2$ error normalized by the image size. We remove the background as in \cite{zhang2018unsupervised, lorenz2019unsupervised} to make a fair comparison to others.
To underline the robustness against structured backgrounds we also report the numbers including background.

\noindent\parag{DeepFashion} \cite{liu2016deepfashion} contains 53k in-shop clothes images. We follow \cite{lorenz2019unsupervised} only keeping the full-body images. We use 10604 images for training and 1179 images for testing as in \cite{sandro2020unsupervised}. We use the keypoints generated by AlphaPose \cite{fang2017rmpe} as the ground truth. The evaluation metric is Percentage of Correct Keypoints of d=6 pixels in resolution $256\times 256$.

\noindent\parag{Taichi} \cite{Siarohin_2019_NeurIPS} contains 3049 training videos and 285 test
videos of people performing Tai-Chi, with the various appearance of foreground and background. We follow \cite{siarohin2021motion} using 5000 and 300 images (not contained in training data) for training a linear regression and for testing, respectively. The evaluation metric, mean average error (MAE), is calculated as the sum of the $L_2$ error in resolution $256\times 256$.

\noindent\parag{CUB-200-2011} \cite{WahCUB_200_2011} consists of 11,788 images of birds. We follow two established protocols \cite{lorenz2019unsupervised, choudhury2021unsupervised} to evaluate our method: 1) Images are cropped based on the bird landmarks, aligned to face to the left \cite{lorenz2019unsupervised}, and seabirds are removed; 2) Birds are cropped based on the given bounding box and the train/val/test split of \cite{choudhury2021unsupervised} is used. In both cases, the evaluation metric is the regressed (without bias) mean $L_2$ error normalized by the cropped image size.

\noindent\parag{Flower} \cite{Nilsback08}, \textbf{11k Hands} \cite{afifi201911kHands}, \textbf{Horses} \cite{zhu2017unpaired}, and \textbf{Zebras} \cite{zhu2017unpaired} are used for qualitative experiments. \textbf{VoxCeleb2} \cite{chung18voxceleb2} and \textbf{AFHQ} \cite{choi2020stargan} are used for pose transfer and conditional image generation, respectively. 
Horses and Zebras are extracted from the CycleGAN dataset \cite{zhu2017unpaired} by removing the images with multiple horses and aligning them to face left. Note that the horses and zebra are trained separately, yet the model learns similar structures. The train/test split of Flower follows \cite{chen2019unsupervised}. All the other datasets follow the train/test split specified by the dataset.

\subsection{Qualitative Analysis}

We qualitatively compare our detected keypoints with other methods and show the examples of the learned edges in Figure~\ref{fig:qualitative}. For visualization purposes, we scale the edge weights 
to obtain visible edges.
We use the same number of keypoints as the previous method \cite{he2022ganseg} for a fair comparison, which are 8 for CelebA-in-the-Wild, 16 for DeepFashion, and 10 for Taichi. We will discuss more on the choice of the number of keypoints in Ablation Study~\ref{sec:ablation}. As shown in Figure~\ref{fig:qualitative}, our model not only detects consistent keypoints but also learns reasonable edges, such as human skeletons in DeepFashion and Taichi. For example, the feet are clearly connected with the corresponding knees, and there is no edge between the left and right hands. We show 105 images with detected keypoints and visualized graph structure for each dataset in Figure~\ref{fig:gen_afhq_k32}-\ref{fig:taichi_k32} in the Appendix, demonstrating that our model works on various classes of objects of diverse appearance and complex backgrounds.

\subsection{Quantitative Analysis} \label{sec:quantitative}

We compare the keypoint detection results with other methods in Table~\ref{tab:celeba} (CelebA), Table~\ref{tab:human_body} (Human3.6, DeepFashion, Taichi), and Table~\ref{tab:cub} (CUB).
Our simple model outperforms all other unsupervised methods in all benchmarks, except \cite{hung2019scops} on three CUB subsets, which however requires saliency maps, and for the most constrained setting CelebA-aligned and CUB-aligned where all methods perform well and the results are comparable. 
The results on CelebA in Table~\ref{tab:celeba} confirms that our model is more robust to poses in the wild. Since self-supervised part segmentation methods are usually more robust on wild faces \cite{liu2021unsupervised, hung2019scops, he2022ganseg}, we also include them for comparison, demonstrating the robustness of our model over existing baselines.
The results on Human3.6m and DeepFashion in Table~\ref{tab:human_body} show the capability of our model to detect keypoints on human bodies of either similar or diverse appearances.
The result on Taichi in Table~\ref{tab:human_body} demonstrates the general applicability of our model to human bodies of diverse poses in complex backgrounds.

\begin{figure}[t]
\begin{center}
  \includegraphics[width=1\textwidth]{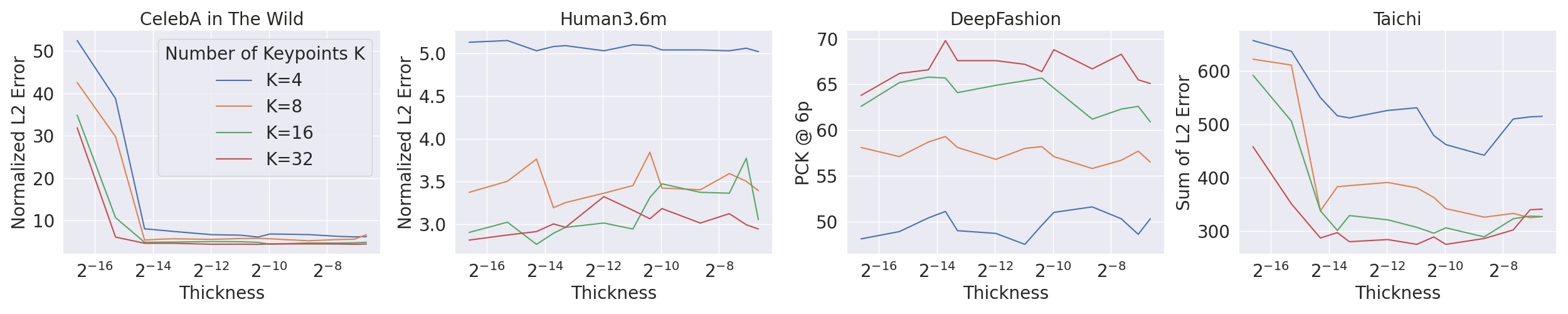}
\end{center}
   \caption{\textbf{Ablation tests on the number of keypoints and edge thickness.} While the model shows better performance with more keypoints, it is robust to the edge thickness.}
\label{fig:ablation_kp_thick}
\end{figure}

\begin{figure}[t]
\begin{center}
  \includegraphics[width=1\textwidth]{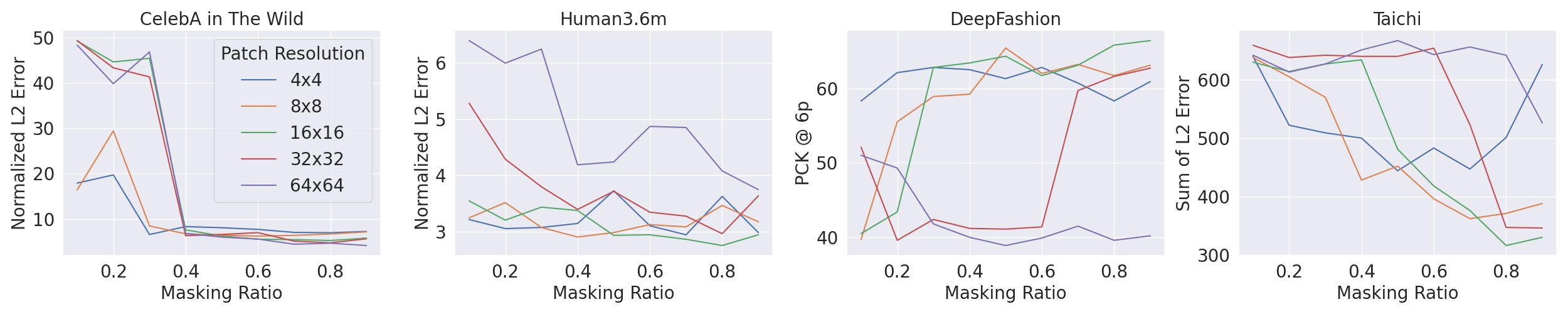}
\end{center}
   \caption{\textbf{Ablation tests on masking strategy.} Overall, the performance increases as the mask ratio increases. Too small or too large patch sizes can decrease the performance. Empirically, masking 80\% of the $16\times16$ patches is a golden rule.}
\label{fig:ablation_masking}
\end{figure}

\subsection{Ablation Tests} \label{sec:ablation}

In this section, we analyze the hyperparameters and demonstrate the robustness of our model. We also discuss the possible variants of our model and show the superiority of our design. 

\begin{figure}[t]
\begin{center}
  \includegraphics[width=1\textwidth]{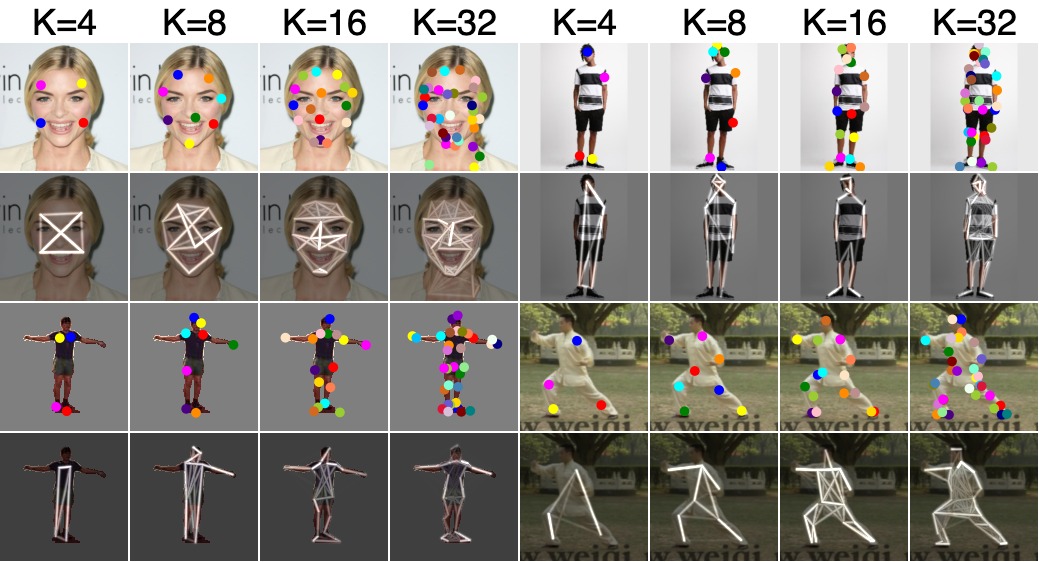}
\end{center}
   \caption{\textbf{Examples of different numbers of detected keypoints.} With very few keypoints, the model only models a very basic shape, such as the box on the face, and sometimes cannot fully capture the structure. For example, with K=4, only legs are modeled on humans. With abundant keypoints, it is able to model details.}
\label{fig:ablation_kp}
\end{figure}


\parag{Number of Keypoints \& Thickness}. 
We show ablation test results on the different numbers of keypoints and edge thicknesses in Figure~\ref{fig:ablation_kp_thick}. The exact numbers can be found in Table~\ref{tab:hyper} in Appendix~\ref{supp:ablation_test}. Our model shows very strong robustness to edge thickness. The accuracy remains state-of-the-art while the thickness of the edges varies by multiple orders of magnitude.
On the other hand, with the increasing number of keypoints the accuracy increases. This is expected since more keypoints are able to capture structure in more detail, as shown in Figure~\ref{fig:ablation_kp}. Yet, some other methods fail for a large number of keypoints \cite{he2022ganseg}.

\parag{Masking Strategy}. 
In our standard setting, the image is divided into $16\times 16$ patches and 80\% of the patches are randomly masked. We investigate how the patch size and masking ratio affect the model performance. Figure~\ref{fig:ablation_masking} shows that a too low masking ratio enables the network to directly infer the structure from the masked image which is undesired in our case.
In these cases, the network would not choose to infer a set of compact keypoints from the original image. Figure~\ref{fig:ablation_masking} illustrates that the patch size cannot be too small ($4\times 4$) or too large ($64\times 64$). Although in some cases, such as CelebA-in-the-Wild, a different masking strategy gives better results (4.14 vs 5.24), we choose to report a unified strategy that we mask 80\% of $16\times16$ patches for simplicity and conciseness.

\begin{table}[t]
\centering
\caption{\textbf{Ablation tests on variants of edge heatmap generation}. The original design is proved to be the most robust one. Although in some cases it is not optimal, the difference is almost trivial.}
\resizebox{0.9\linewidth}{!}{%
\begin{tabular}{lcccc}
\Xhline{1.5pt}
Model & CelebA in The Wild $\downarrow$ & Human3.6m $\downarrow$ & DeepFashion $\uparrow$ & Taichi $\downarrow$\\ \Xhline{1.5pt}
original model & \textbf{5.24}\%  & \textbf{2.76}\% & 65.8\% & 316 \\ 
fixed $\alpha$ & 6.39\%  & 2.87\% & \textbf{66.0}\% & 374 \\
shared learnable thickness & 6.12\%  & 3.25\% & 49.1\% & 425 \\
independent learnable thickness & 5.94\%  & 3.73\% & 50.2\% & \textbf{311} \\
edge-specific heatmap & 5.65\%  & 3.83\% & 65.1\% & 407\\
only using keypoints without edges & 6.55\% & 3.58\% & 52.9\% & 722\\
\Xhline{1.5pt}
\end{tabular}
}
\label{tab:ablation_edge_gen}
\end{table}

\begin{figure}[t]
\begin{center}
  \includegraphics[width=1.\textwidth]{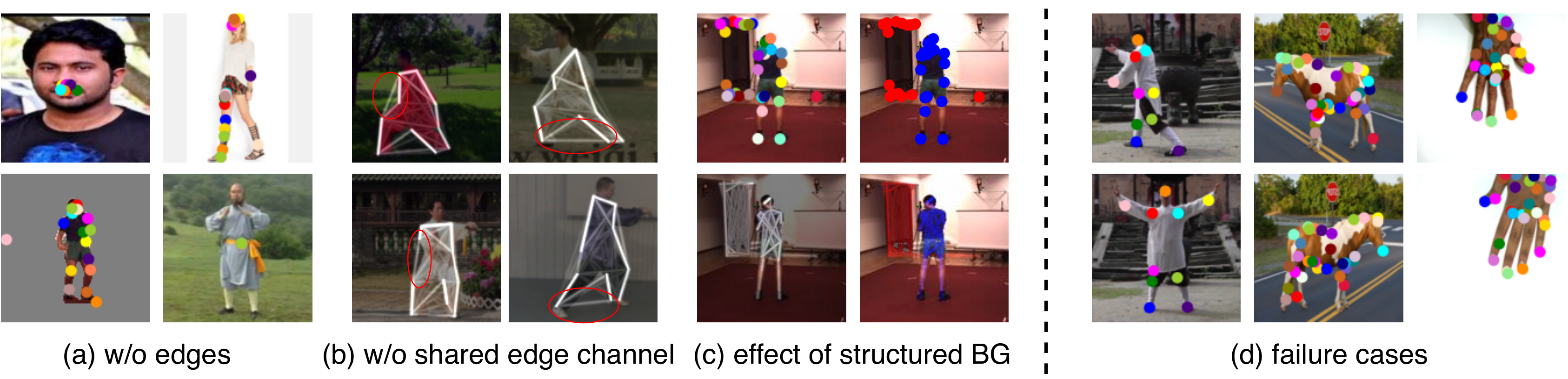}
\end{center}
   \caption{(a) If we do not model the edges, the model may degenerate. (b) If we give different edges a different channel in the feature map, the model would give dummy edges. (c) If the model is trained on the dataset with structured background, the background would be modeled. However, the keypoints can be separated into two sets by spectral clustering. (d) failure case: left) the model cannot model the occlusion well; right) the model has left and right ambiguity.}
\label{fig:ablation_edge}
\end{figure}

\parag{Variants of Edge Heatmap Generation}. 
Besides the heatmap generation described in Method~\ref{sec:method}, we test four more ways to generate the edge maps: 1) we define the thickness as a globally learnable parameter; 2) we learn each edge thickness independently as a parameter; 3) we treat each edge heatmap as an independent channel of the feature map, instead of making them a single channel heatmap with Equation~\ref{eq:max_heatmap}; 4) instead of using the edge heatmap, we generate Gaussian heatmaps for the keypoints and use Equation~\ref{eq:max_heatmap} to combine them in a single channel heatmap. The results are listed in Table~\ref{tab:ablation_edge_gen}. Overall, these variants have worse performance. Although in some cases, the model has slightly better results on specific datasets, the performance boost does not hold in general. We observe that with only keypoints without edges, the model may degenerate, as shown in Figure~\ref{fig:ablation_edge}a. Interestingly, assigning each edge a different channel performs worse than simply combining all edges into a single channel. We believe it is caused by entangling the edge weights with the convolution kernel weights. As visualized in Figure~\ref{fig:ablation_edge}b, there exist dummy edges that do not model the object structure.
In addition, we tried to remove the learnable $\alpha$ in Equation~\ref{eq:recon}, fixing $\alpha=1$, but the overall performance decreases as shown in Table~\ref{tab:ablation_edge_gen}.


\parag{Does Texture Matter?} 
We trained two networks on horses and zebras separately. As shown in Figure~\ref{fig:teaser}a, the horse and the zebra share similar shape structures but only one is textured. The striped texture not having a significant impact on the learned structure shows that our model primarily learns the structure instead of texture features.

\parag{What if the model is trained on images with a structured background?} We tested on Human3.6m with a background, where all images are taken in a single room. The error is 5.02. As shown in Figure~\ref{fig:ablation_edge}c, our model captures the entrance in the background. It is expected since we assume the foreground object is structured. If we apply spectral clustering \cite{ng2001spectral} on the learned graph, the keypoints are clearly divided into two clusters, one for the room and one for the person.
\section{Limitations and Future Work} \label{sec:limitation}
If the background is highly structured, the keypoints will appear on the background. Yet, 
we showed an avenue for future work, as already a simple 
graph clustering could separate the object from the background on the Human3.6M dataset. Similar to the previous 2D self-supervised methods \cite{lorenz2019unsupervised, siarohin2021motion, he2022ganseg}, our model cannot model occlusion well. We show in Figure~\ref{fig:ablation_edge}d left that the occluded right arm becomes the back when the person turns to the left. In addition, as for all other methods, the model cannot distinguish the left and right sides of the objects as shown in Figure~\ref{fig:ablation_edge}d middle and right. We believe it is necessary to model the structure in 3D to solve these problems.


\section{Conclusion}

We presented a simple approach for learning a spatial graph representation from unlabelled image collections by reconstructing masked images. The crucial part is our learnable graph design that models the relationship between different keypoints. 
It is simpler than existing alternatives 
and opens up a path for image understanding, image editing, and learning 3D models from 2D images.

\section*{Acknowledgement}
This work was supported by the Compute Canada GPU servers, and a Huawei-UBC Joint Lab project.

\bibliographystyle{abbrvnat}
\bibliography{egbib}

\begin{thebibliography}{131}
\providecommand{\natexlab}[1]{#1}
\providecommand{\url}[1]{\texttt{#1}}
\expandafter\ifx\csname urlstyle\endcsname\relax
  \providecommand{\doi}[1]{doi: #1}\else
  \providecommand{\doi}{doi: \begingroup \urlstyle{rm}\Url}\fi

\bibitem[Afifi(2019)]{afifi201911kHands}
M.~Afifi.
\newblock 11k hands: gender recognition and biometric identification using a
  large dataset of hand images.
\newblock \emph{Multimedia Tools and Applications}, 2019.
\newblock \doi{10.1007/s11042-019-7424-8}.
\newblock URL \url{https://doi.org/10.1007/s11042-019-7424-8}.

\bibitem[Andriluka et~al.(2014)Andriluka, Pishchulin, Gehler, and
  Schiele]{andriluka14cvpr}
M.~Andriluka, L.~Pishchulin, P.~Gehler, and B.~Schiele.
\newblock 2d human pose estimation: New benchmark and state of the art
  analysis.
\newblock In \emph{IEEE Conference on Computer Vision and Pattern Recognition
  (CVPR)}, June 2014.

\bibitem[Andriluka et~al.(2018)Andriluka, Iqbal, Insafutdinov, Pishchulin,
  Milan, Gall, and B.]{PoseTrack}
M.~Andriluka, U.~Iqbal, E.~Insafutdinov, L.~Pishchulin, A.~Milan, J.~Gall, and
  S.~B.
\newblock Pose{T}rack: {A} benchmark for human pose estimation and tracking.
\newblock In \emph{CVPR}, 2018.

\bibitem[Arandjelovi{\'c} and Zisserman(2019)]{arandjelovic2019object}
R.~Arandjelovi{\'c} and A.~Zisserman.
\newblock Object discovery with a copy-pasting gan.
\newblock \emph{arXiv preprint arXiv:1905.11369}, 2019.

\bibitem[Blanz and Vetter(1999)]{blanz1999morphable}
V.~Blanz and T.~Vetter.
\newblock A morphable model for the synthesis of 3d faces.
\newblock In \emph{Proceedings of the 26th annual conference on Computer
  graphics and interactive techniques}, pages 187--194, 1999.

\bibitem[Braun(2020)]{sandro2020unsupervised}
S.~Braun.
\newblock unsupervised-disentangling.
\newblock
  \url{https://github.com/theRealSuperMario/unsupervised-disentangling/tree/reproducing_baselines},
  2020.

\bibitem[Cao and Fu(2021)]{cao2021learning}
C.~Cao and Y.~Fu.
\newblock Learning a sketch tensor space for image inpainting of man-made
  scenes.
\newblock In \emph{Proceedings of the IEEE/CVF International Conference on
  Computer Vision}, pages 14509--14518, 2021.

\bibitem[Cao et~al.(2019)Cao, Yan, Shi, and Chen]{Cao2019ai}
N.~Cao, X.~Yan, Y.~Shi, and C.~Chen.
\newblock Ai-sketcher : A deep generative model for producing high-quality
  sketches.
\newblock \emph{Proceedings of the AAAI Conference on Artificial Intelligence},
  33\penalty0 (01):\penalty0 2564--2571, Jul. 2019.
\newblock \doi{10.1609/aaai.v33i01.33012564}.
\newblock URL \url{https://ojs.aaai.org/index.php/AAAI/article/view/4103}.

\bibitem[Cao et~al.(2017)Cao, Simon, Wei, and Sheikh]{cao2017realtime}
Z.~Cao, T.~Simon, S.-E. Wei, and Y.~Sheikh.
\newblock Realtime multi-person 2d pose estimation using part affinity fields.
\newblock In \emph{Proceedings of the IEEE conference on computer vision and
  pattern recognition}, pages 7291--7299, 2017.

\bibitem[Chen et~al.(2019)Chen, Arti{\`e}res, and
  Denoyer]{chen2019unsupervised}
M.~Chen, T.~Arti{\`e}res, and L.~Denoyer.
\newblock Unsupervised object segmentation by redrawing.
\newblock In \emph{Advances in Neural Information Processing Systems 32 (NIPS
  2019)}, pages 12705--12716, 2019.

\bibitem[Chen et~al.(2020)Chen, Kornblith, Norouzi, and Hinton]{chen2020simple}
T.~Chen, S.~Kornblith, M.~Norouzi, and G.~Hinton.
\newblock A simple framework for contrastive learning of visual
  representations.
\newblock In \emph{International conference on machine learning}, pages
  1597--1607. PMLR, 2020.

\bibitem[Chen et~al.(2017)Chen, Tu, Yi, and Xu]{chen2017sketch}
Y.~Chen, S.~Tu, Y.~Yi, and L.~Xu.
\newblock Sketch-pix2seq: a model to generate sketches of multiple categories.
\newblock \emph{arXiv preprint arXiv:1709.04121}, 2017.

\bibitem[Cheng et~al.(2011)Cheng, Zhang, Mitra, Huang, and Hu]{cheng2011global}
M.-M. Cheng, G.-X. Zhang, N.~J. Mitra, X.~Huang, and S.-M. Hu.
\newblock Global contrast based salient region detection.
\newblock In \emph{CVPR 2011}, pages 409--416, 2011.
\newblock \doi{10.1109/CVPR.2011.5995344}.

\bibitem[Choi et~al.(2020)Choi, Uh, Yoo, and Ha]{choi2020stargan}
Y.~Choi, Y.~Uh, J.~Yoo, and J.-W. Ha.
\newblock Stargan v2: Diverse image synthesis for multiple domains.
\newblock In \emph{Proceedings of the IEEE/CVF conference on computer vision
  and pattern recognition}, pages 8188--8197, 2020.

\bibitem[Choudhury et~al.(2021)Choudhury, Laina, Rupprecht, and
  Vedaldi]{choudhury2021unsupervised}
S.~Choudhury, I.~Laina, C.~Rupprecht, and A.~Vedaldi.
\newblock Unsupervised part discovery from contrastive reconstruction.
\newblock In A.~Beygelzimer, Y.~Dauphin, P.~Liang, and J.~W. Vaughan, editors,
  \emph{Advances in Neural Information Processing Systems}, 2021.
\newblock URL \url{https://openreview.net/forum?id=iHXQPrISusS}.

\bibitem[Chung et~al.(2018)Chung, Nagrani, and Zisserman]{chung18voxceleb2}
J.~S. Chung, A.~Nagrani, and A.~Zisserman.
\newblock Voxceleb2: Deep speaker recognition.
\newblock In \emph{INTERSPEECH}, 2018.

\bibitem[Collins et~al.(2018)Collins, Achanta, and Susstrunk]{collins2018deep}
E.~Collins, R.~Achanta, and S.~Susstrunk.
\newblock Deep feature factorization for concept discovery.
\newblock In \emph{Proceedings of the European Conference on Computer Vision
  (ECCV)}, pages 336--352, 2018.

\bibitem[Crawford and Pineau(2019)]{crawford2019spatially}
E.~Crawford and J.~Pineau.
\newblock Spatially invariant unsupervised object detection with convolutional
  neural networks.
\newblock In \emph{Proceedings of the AAAI Conference on Artificial
  Intelligence}, pages 3412--3420, 2019.

\bibitem[Das et~al.(2020)Das, Yang, Hospedales, Xiang, and
  Song]{das2020beziersketch}
A.~Das, Y.~Yang, T.~Hospedales, T.~Xiang, and Y.-Z. Song.
\newblock B{\'e}ziersketch: A generative model for scalable vector sketches.
\newblock In \emph{European Conference on Computer Vision}, pages 632--647.
  Springer, 2020.

\bibitem[Dong et~al.(2018)Dong, Yu, Weng, Wei, Yang, and
  Sheikh]{dong2018supervision}
X.~Dong, S.-I. Yu, X.~Weng, S.-E. Wei, Y.~Yang, and Y.~Sheikh.
\newblock Supervision-by-registration: An unsupervised approach to improve the
  precision of facial landmark detectors.
\newblock In \emph{Proceedings of the IEEE Conference on Computer Vision and
  Pattern Recognition}, pages 360--368, 2018.

\bibitem[Dosovitskiy and Brox(2016)]{dosovitskiy2016generating}
A.~Dosovitskiy and T.~Brox.
\newblock Generating images with perceptual similarity metrics based on deep
  networks.
\newblock \emph{Advances in neural information processing systems}, 29, 2016.

\bibitem[Dugas et~al.(2000)Dugas, Bengio, B\'{e}lisle, Nadeau, and
  Garcia]{dugas2000incorporating}
C.~Dugas, Y.~Bengio, F.~B\'{e}lisle, C.~Nadeau, and R.~Garcia.
\newblock Incorporating second-order functional knowledge for better option
  pricing.
\newblock In T.~Leen, T.~Dietterich, and V.~Tresp, editors, \emph{Advances in
  Neural Information Processing Systems}, volume~13. MIT Press, 2000.
\newblock URL
  \url{https://proceedings.neurips.cc/paper/2000/file/44968aece94f667e4095002d140b5896-Paper.pdf}.

\bibitem[Dundar et~al.(2020)Dundar, Shih, Garg, Pottorf, Tao, and
  Catanzaro]{dundar2020unsupervised}
A.~Dundar, K.~J. Shih, A.~Garg, R.~Pottorf, A.~Tao, and B.~Catanzaro.
\newblock Unsupervised disentanglement of pose, appearance and background from
  images and videos.
\newblock \emph{arXiv preprint arXiv:2001.09518}, 2020.

\bibitem[Eitz et~al.(2012)Eitz, Hays, and Alexa]{eitz2012hdhso}
M.~Eitz, J.~Hays, and M.~Alexa.
\newblock How do humans sketch objects?
\newblock \emph{ACM Trans. Graph. (Proc. SIGGRAPH)}, 31\penalty0 (4):\penalty0
  44:1--44:10, 2012.

\bibitem[Fang et~al.(2017)Fang, Xie, Tai, and Lu]{fang2017rmpe}
H.-S. Fang, S.~Xie, Y.-W. Tai, and C.~Lu.
\newblock {RMPE}: Regional multi-person pose estimation.
\newblock In \emph{ICCV}, 2017.

\bibitem[Feng et~al.(2018)Feng, Wu, Shao, Wang, and Zhou]{feng2018joint}
Y.~Feng, F.~Wu, X.~Shao, Y.~Wang, and X.~Zhou.
\newblock Joint 3d face reconstruction and dense alignment with position map
  regression network.
\newblock In \emph{Proceedings of the European conference on computer vision
  (ECCV)}, pages 534--551, 2018.

\bibitem[Gatys et~al.(2016)Gatys, Ecker, and Bethge]{gatys2016image}
L.~A. Gatys, A.~S. Ecker, and M.~Bethge.
\newblock Image style transfer using convolutional neural networks.
\newblock In \emph{Proceedings of the IEEE conference on computer vision and
  pattern recognition}, pages 2414--2423, 2016.

\bibitem[Ge et~al.(2021)Ge, Goswami, Zitnick, and Parikh]{ge2021creative}
S.~Ge, V.~Goswami, L.~Zitnick, and D.~Parikh.
\newblock Creative sketch generation.
\newblock In \emph{International Conference on Learning Representations}, 2021.
\newblock URL \url{https://openreview.net/forum?id=gwnoVHIES05}.

\bibitem[Goodfellow et~al.(2014)Goodfellow, Pouget-Abadie, Mirza, Xu,
  Warde-Farley, Ozair, Courville, and Bengio]{goodfellow2014generative}
I.~Goodfellow, J.~Pouget-Abadie, M.~Mirza, B.~Xu, D.~Warde-Farley, S.~Ozair,
  A.~Courville, and Y.~Bengio.
\newblock Generative adversarial nets.
\newblock In \emph{Advances in neural information processing systems}, pages
  2672--2680, 2014.

\bibitem[Ha and Eck(2018)]{ha2018a}
D.~Ha and D.~Eck.
\newblock A neural representation of sketch drawings.
\newblock In \emph{International Conference on Learning Representations}, 2018.
\newblock URL \url{https://openreview.net/forum?id=Hy6GHpkCW}.

\bibitem[Han et~al.(2019)Han, Wu, Huang, Scott, and Davis]{han2019finet}
X.~Han, Z.~Wu, W.~Huang, M.~R. Scott, and L.~S. Davis.
\newblock Finet: Compatible and diverse fashion image inpainting.
\newblock In \emph{Proceedings of the IEEE/CVF International Conference on
  Computer Vision (ICCV)}, October 2019.

\bibitem[He et~al.(2020)He, Fan, Wu, Xie, and Girshick]{he2020momentum}
K.~He, H.~Fan, Y.~Wu, S.~Xie, and R.~Girshick.
\newblock Momentum contrast for unsupervised visual representation learning.
\newblock In \emph{Proceedings of the IEEE/CVF conference on computer vision
  and pattern recognition}, pages 9729--9738, 2020.

\bibitem[He et~al.(2021{\natexlab{a}})He, Chen, Xie, Li, Doll{\'a}r, and
  Girshick]{he2021masked}
K.~He, X.~Chen, S.~Xie, Y.~Li, P.~Doll{\'a}r, and R.~Girshick.
\newblock Masked autoencoders are scalable vision learners.
\newblock \emph{arXiv preprint arXiv:2111.06377}, 2021{\natexlab{a}}.

\bibitem[He et~al.(2021{\natexlab{b}})He, Wandt, and
  Rhodin]{he2021latentkeypointgan}
X.~He, B.~Wandt, and H.~Rhodin.
\newblock Latentkeypointgan: Controlling gans via latent keypoints.
\newblock \emph{arXiv preprint arXiv:2103.15812}, 2021{\natexlab{b}}.

\bibitem[He et~al.(2022)He, Wandt, and Rhodin]{he2022ganseg}
X.~He, B.~Wandt, and H.~Rhodin.
\newblock Ganseg: Learning to segment by unsupervised hierarchical image
  generation.
\newblock In \emph{Proceedings of the IEEE/CVF Conference on Computer Vision
  and Pattern Recognition}, pages 1225--1235, 2022.

\bibitem[Huang and Li(2020)]{huang2020interpretable}
Z.~Huang and Y.~Li.
\newblock Interpretable and accurate fine-grained recognition via region
  grouping.
\newblock In \emph{Proceedings of the IEEE/CVF Conference on Computer Vision
  and Pattern Recognition}, pages 8662--8672, 2020.

\bibitem[Hung et~al.(2019)Hung, Jampani, Liu, Molchanov, Yang, and
  Kautz]{hung2019scops}
W.-C. Hung, V.~Jampani, S.~Liu, P.~Molchanov, M.-H. Yang, and J.~Kautz.
\newblock Scops: Self-supervised co-part segmentation.
\newblock In \emph{Proceedings of the IEEE/CVF Conference on Computer Vision
  and Pattern Recognition}, pages 869--878, 2019.

\bibitem[Ioffe and Szegedy(2015)]{ioffe2015batch}
S.~Ioffe and C.~Szegedy.
\newblock Batch normalization: Accelerating deep network training by reducing
  internal covariate shift.
\newblock \emph{arXiv preprint arXiv:1502.03167}, 2015.

\bibitem[Ionescu et~al.(2014)Ionescu, Papava, Olaru, and
  Sminchisescu]{Ionescu2014human36m}
C.~Ionescu, D.~Papava, V.~Olaru, and C.~Sminchisescu.
\newblock Human3.6m: Large scale datasets and predictive methods for 3d human
  sensing in natural environments.
\newblock \emph{IEEE Transactions on Pattern Analysis and Machine
  Intelligence}, 36\penalty0 (7):\penalty0 1325--1339, 2014.
\newblock \doi{10.1109/TPAMI.2013.248}.

\bibitem[Isola et~al.(2017)Isola, Zhu, Zhou, and Efros]{isola2017image}
P.~Isola, J.-Y. Zhu, T.~Zhou, and A.~A. Efros.
\newblock Image-to-image translation with conditional adversarial networks.
\newblock In \emph{Proceedings of the IEEE conference on computer vision and
  pattern recognition}, pages 1125--1134, 2017.

\bibitem[Jackson et~al.(2017)Jackson, Bulat, Argyriou, and
  Tzimiropoulos]{jackson2017large}
A.~S. Jackson, A.~Bulat, V.~Argyriou, and G.~Tzimiropoulos.
\newblock Large pose 3d face reconstruction from a single image via direct
  volumetric cnn regression.
\newblock In \emph{Proceedings of the IEEE international conference on computer
  vision}, pages 1031--1039, 2017.

\bibitem[Jakab et~al.(2018)Jakab, Gupta, Bilen, and
  Vedaldi]{jakab2018unsupervised}
T.~Jakab, A.~Gupta, H.~Bilen, and A.~Vedaldi.
\newblock Unsupervised learning of object landmarks through conditional image
  generation.
\newblock In \emph{Advances in neural information processing systems}, pages
  4016--4027, 2018.

\bibitem[Jakab et~al.(2020)Jakab, Gupta, Bilen, and Vedaldi]{jakab2020self}
T.~Jakab, A.~Gupta, H.~Bilen, and A.~Vedaldi.
\newblock Self-supervised learning of interpretable keypoints from unlabelled
  videos.
\newblock In \emph{Proceedings of the IEEE/CVF Conference on Computer Vision
  and Pattern Recognition}, pages 8787--8797, 2020.

\bibitem[Jiang et~al.(2013)Jiang, Wang, Yuan, Wu, Zheng, and
  Li]{jiang2013salient}
H.~Jiang, J.~Wang, Z.~Yuan, Y.~Wu, N.~Zheng, and S.~Li.
\newblock Salient object detection: A discriminative regional feature
  integration approach.
\newblock In \emph{2013 IEEE Conference on Computer Vision and Pattern
  Recognition}, pages 2083--2090, 2013.
\newblock \doi{10.1109/CVPR.2013.271}.

\bibitem[Jie~Yang(2020)]{jie2020inpainting}
Y.~S. Jie~Yang, Zhiquan~Qi.
\newblock Learning to incorporate structure knowledge for image inpainting.
\newblock In \emph{Proceedings of the AAAI Conference on Artificial
  Intelligence}, pages 12605--12612, 2020.

\bibitem[Johnson et~al.(2016)Johnson, Alahi, and
  Fei-Fei]{johnson2016perceptual}
J.~Johnson, A.~Alahi, and L.~Fei-Fei.
\newblock Perceptual losses for real-time style transfer and super-resolution.
\newblock In \emph{European conference on computer vision}, pages 694--711.
  Springer, 2016.

\bibitem[Kaiyrbekov and Sezgin(2019)]{kaiyrbekov2019deep}
K.~Kaiyrbekov and M.~Sezgin.
\newblock Deep stroke-based sketched symbol reconstruction and segmentation.
\newblock \emph{IEEE computer graphics and applications}, 40\penalty0
  (1):\penalty0 112--126, 2019.

\bibitem[Kampelmuhler and Pinz(2020)]{kampelmuhler2020synthesizing}
M.~Kampelmuhler and A.~Pinz.
\newblock Synthesizing human-like sketches from natural images using a
  conditional convolutional decoder.
\newblock In \emph{Proceedings of the IEEE/CVF Winter Conference on
  Applications of Computer Vision}, pages 3203--3211, 2020.

\bibitem[Karras et~al.(2019)Karras, Laine, and Aila]{karras2019style}
T.~Karras, S.~Laine, and T.~Aila.
\newblock A style-based generator architecture for generative adversarial
  networks.
\newblock In \emph{Proceedings of the IEEE conference on computer vision and
  pattern recognition}, pages 4401--4410, 2019.

\bibitem[Karras et~al.(2020)Karras, Laine, Aittala, Hellsten, Lehtinen, and
  Aila]{karras2020analyzing}
T.~Karras, S.~Laine, M.~Aittala, J.~Hellsten, J.~Lehtinen, and T.~Aila.
\newblock Analyzing and improving the image quality of stylegan.
\newblock In \emph{Proceedings of the IEEE/CVF Conference on Computer Vision
  and Pattern Recognition}, pages 8110--8119, 2020.

\bibitem[Katircioglu et~al.(2021{\natexlab{a}})Katircioglu, Rhodin, Constantin,
  Sporri, Salzmann, and Fua]{katircioglu2021self}
I.~Katircioglu, H.~Rhodin, V.~Constantin, J.~Sporri, M.~Salzmann, and P.~Fua.
\newblock Self-supervised human detection and segmentation via background
  inpainting.
\newblock \emph{IEEE Transactions on Pattern Analysis \& Machine Intelligence},
  01:\penalty0 1--1, 2021{\natexlab{a}}.

\bibitem[Katircioglu et~al.(2021{\natexlab{b}})Katircioglu, Rhodin, Sporri,
  Salzmann, and Fua]{katircioglu2021human}
I.~Katircioglu, H.~Rhodin, J.~Sporri, M.~Salzmann, and P.~Fua.
\newblock Human detection and segmentation via multi-view consensus.
\newblock In \emph{Proceedings of the IEEE/CVF International Conference on
  Computer Vision}, pages 2855--2864, 2021{\natexlab{b}}.

\bibitem[Kim et~al.(2019)Kim, Nam, Cho, and Kim]{kim2019unsupervised}
Y.~Kim, S.~Nam, I.~Cho, and S.~J. Kim.
\newblock Unsupervised keypoint learning for guiding class-conditional video
  prediction.
\newblock In \emph{Advances in Neural Information Processing Systems}, pages
  3814--3824, 2019.

\bibitem[Kingma and Ba(2015)]{KingmaB14}
D.~P. Kingma and J.~Ba.
\newblock Adam: A method for stochastic optimization.
\newblock In \emph{International Conference on Learning Representations}, 2015.

\bibitem[Kulkarni et~al.(2019)Kulkarni, Gupta, Ionescu, Borgeaud, Reynolds,
  Zisserman, and Mnih]{kulkarni2019unsupervised}
T.~D. Kulkarni, A.~Gupta, C.~Ionescu, S.~Borgeaud, M.~Reynolds, A.~Zisserman,
  and V.~Mnih.
\newblock Unsupervised learning of object keypoints for perception and control.
\newblock In \emph{Advances in neural information processing systems}, pages
  10724--10734, 2019.

\bibitem[Lahiri et~al.(2020)Lahiri, Jain, Agrawal, Mitra, and
  Biswas]{lahiri2020prior}
A.~Lahiri, A.~K. Jain, S.~Agrawal, P.~Mitra, and P.~K. Biswas.
\newblock Prior guided gan based semantic inpainting.
\newblock In \emph{Proceedings of the IEEE/CVF Conference on Computer Vision
  and Pattern Recognition (CVPR)}, June 2020.

\bibitem[LeCun et~al.(1989)LeCun, Boser, Denker, Henderson, Howard, Hubbard,
  and Jackel]{lecun1989backpropagation}
Y.~LeCun, B.~Boser, J.~S. Denker, D.~Henderson, R.~E. Howard, W.~Hubbard, and
  L.~D. Jackel.
\newblock Backpropagation applied to handwritten zip code recognition.
\newblock \emph{Neural Computation}, 1\penalty0 (4):\penalty0 541--551, 1989.
\newblock \doi{10.1162/neco.1989.1.4.541}.

\bibitem[Li et~al.(2019)Li, He, Zhang, Du, and Tao]{li2019progressive}
J.~Li, F.~He, L.~Zhang, B.~Du, and D.~Tao.
\newblock Progressive reconstruction of visual structure for image inpainting.
\newblock In \emph{Proceedings of the IEEE/CVF International Conference on
  Computer Vision (ICCV)}, October 2019.

\bibitem[Li et~al.(2017)Li, Bolkart, Black, Li, and
  Romero]{FLAME:SiggraphAsia2017}
T.~Li, T.~Bolkart, M.~J. Black, H.~Li, and J.~Romero.
\newblock Learning a model of facial shape and expression from {4D} scans.
\newblock \emph{ACM Transactions on Graphics, (Proc. SIGGRAPH Asia)},
  36\penalty0 (6):\penalty0 194:1--194:17, 2017.

\bibitem[Liao et~al.(2020)Liao, Xiao, Wang, Lin, and Satoh]{liao2020guidance}
L.~Liao, J.~Xiao, Z.~Wang, C.-W. Lin, and S.~Satoh.
\newblock Guidance and evaluation: Semantic-aware image inpainting for mixed
  scenes.
\newblock In \emph{European Conference on Computer Vision}, pages 683--700.
  Springer, 2020.

\bibitem[Lin et~al.(2014)Lin, Maire, Belongie, Hays, Perona, Ramanan,
  Doll{\'a}r, and Zitnick]{lin2014microsoft}
T.-Y. Lin, M.~Maire, S.~Belongie, J.~Hays, P.~Perona, D.~Ramanan,
  P.~Doll{\'a}r, and C.~L. Zitnick.
\newblock Microsoft coco: Common objects in context.
\newblock In \emph{European conference on computer vision}, pages 740--755.
  Springer, 2014.

\bibitem[Liu et~al.(2021)Liu, Zhang, Yang, Su, and Zhu]{liu2021unsupervised}
S.~Liu, L.~Zhang, X.~Yang, H.~Su, and J.~Zhu.
\newblock Unsupervised part segmentation through disentangling appearance and
  shape.
\newblock In \emph{Proceedings of the IEEE/CVF Conference on Computer Vision
  and Pattern Recognition}, pages 8355--8364, 2021.

\bibitem[Liu et~al.(2015)Liu, Luo, Wang, and Tang]{liu2015faceattributes}
Z.~Liu, P.~Luo, X.~Wang, and X.~Tang.
\newblock Deep learning face attributes in the wild.
\newblock In \emph{Proceedings of International Conference on Computer Vision
  (ICCV)}, December 2015.

\bibitem[Liu et~al.(2016)Liu, Luo, Qiu, Wang, and Tang]{liu2016deepfashion}
Z.~Liu, P.~Luo, S.~Qiu, X.~Wang, and X.~Tang.
\newblock Deepfashion: Powering robust clothes recognition and retrieval with
  rich annotations.
\newblock In \emph{Proceedings of IEEE Conference on Computer Vision and
  Pattern Recognition (CVPR)}, June 2016.

\bibitem[Loper et~al.(2015)Loper, Mahmood, Romero, Pons-Moll, and
  Black]{SMPL2015}
M.~Loper, N.~Mahmood, J.~Romero, G.~Pons-Moll, and M.~J. Black.
\newblock {SMPL}: A skinned multi-person linear model.
\newblock \emph{ACM Trans. Graphics (Proc. SIGGRAPH Asia)}, 34\penalty0
  (6):\penalty0 248:1--248:16, Oct. 2015.

\bibitem[Lorenz et~al.(2019)Lorenz, Bereska, Milbich, and
  Ommer]{lorenz2019unsupervised}
D.~Lorenz, L.~Bereska, T.~Milbich, and B.~Ommer.
\newblock Unsupervised part-based disentangling of object shape and appearance.
\newblock In \emph{Proceedings of the IEEE Conference on Computer Vision and
  Pattern Recognition}, pages 10955--10964, 2019.

\bibitem[Ma et~al.(2017)Ma, Jia, Sun, Schiele, Tuytelaars, and
  Van~Gool]{ma2017pose}
L.~Ma, X.~Jia, Q.~Sun, B.~Schiele, T.~Tuytelaars, and L.~Van~Gool.
\newblock Pose guided person image generation.
\newblock \emph{Advances in neural information processing systems}, 30, 2017.

\bibitem[Ma et~al.(2018)Ma, Sun, Georgoulis, Van~Gool, Schiele, and
  Fritz]{ma2018disentangled}
L.~Ma, Q.~Sun, S.~Georgoulis, L.~Van~Gool, B.~Schiele, and M.~Fritz.
\newblock Disentangled person image generation.
\newblock In \emph{Proceedings of the IEEE Conference on Computer Vision and
  Pattern Recognition}, pages 99--108, 2018.

\bibitem[Maas et~al.(2013)Maas, Hannun, and Ng]{maas2013rectifier}
A.~L. Maas, A.~Y. Hannun, and A.~Y. Ng.
\newblock Rectifier nonlinearities improve neural network acoustic models.
\newblock In \emph{in ICML Workshop on Deep Learning for Audio, Speech and
  Language Processing}. Citeseer, 2013.

\bibitem[Malladi et~al.(1995)Malladi, Sethian, and Vemuri]{malladi1995shape}
R.~Malladi, J.~Sethian, and B.~Vemuri.
\newblock Shape modeling with front propagation: a level set approach.
\newblock \emph{IEEE Transactions on Pattern Analysis and Machine
  Intelligence}, 17\penalty0 (2):\penalty0 158--175, 1995.
\newblock \doi{10.1109/34.368173}.

\bibitem[Mehta et~al.(2017)Mehta, Rhodin, Casas, Fua, Sotnychenko, Xu, and
  Theobalt]{mono3dhp2017}
D.~Mehta, H.~Rhodin, D.~Casas, P.~Fua, O.~Sotnychenko, W.~Xu, and C.~Theobalt.
\newblock Monocular 3d human pose estimation in the wild using improved cnn
  supervision.
\newblock In \emph{3D Vision (3DV), 2017 Fifth International Conference on}.
  IEEE, 2017.
\newblock \doi{10.1109/3dv.2017.00064}.
\newblock URL \url{http://gvv.mpi-inf.mpg.de/3dhp_dataset}.

\bibitem[Mescheder et~al.(2018)Mescheder, Geiger, and
  Nowozin]{mescheder2018training}
L.~Mescheder, A.~Geiger, and S.~Nowozin.
\newblock Which training methods for gans do actually converge?
\newblock In \emph{International conference on machine learning}, pages
  3481--3490. PMLR, 2018.

\bibitem[Mihai and Hare(2021)]{mihai2021differentiable}
D.~Mihai and J.~Hare.
\newblock Differentiable drawing and sketching.
\newblock \emph{arXiv preprint arXiv:2103.16194}, 2021.

\bibitem[Minderer et~al.(2019)Minderer, Sun, Villegas, Cole, Murphy, and
  Lee]{minderer2019unsupervised}
M.~Minderer, C.~Sun, R.~Villegas, F.~Cole, K.~P. Murphy, and H.~Lee.
\newblock Unsupervised learning of object structure and dynamics from videos.
\newblock In \emph{Advances in Neural Information Processing Systems}, pages
  92--102, 2019.

\bibitem[Nazeri et~al.(2019)Nazeri, Ng, Joseph, Qureshi, and
  Ebrahimi]{nazeri2019edgeconnect}
K.~Nazeri, E.~Ng, T.~Joseph, F.~Qureshi, and M.~Ebrahimi.
\newblock Edgeconnect: Structure guided image inpainting using edge prediction.
\newblock In \emph{The IEEE International Conference on Computer Vision (ICCV)
  Workshops}, Oct 2019.

\bibitem[Ng et~al.(2001)Ng, Jordan, and Weiss]{ng2001spectral}
A.~Y. Ng, M.~I. Jordan, and Y.~Weiss.
\newblock On spectral clustering: Analysis and an algorithm.
\newblock In \emph{Proceedings of the 14th International Conference on Neural
  Information Processing Systems: Natural and Synthetic}, NIPS'01, page
  849–856, Cambridge, MA, USA, 2001. MIT Press.

\bibitem[Nilsback and Zisserman(2008)]{Nilsback08}
M.-E. Nilsback and A.~Zisserman.
\newblock Automated flower classification over a large number of classes.
\newblock In \emph{Indian Conference on Computer Vision, Graphics and Image
  Processing}, Dec 2008.

\bibitem[Noguchi et~al.(2022)Noguchi, Iqbal, Tremblay, Harada, and
  Gallo]{noguchi2021watch}
A.~Noguchi, U.~Iqbal, J.~Tremblay, T.~Harada, and O.~Gallo.
\newblock Watch it move: {U}nsupervised discovery of {3D} joints for re-posing
  of articulated objects.
\newblock In \emph{Proceedings of the IEEE/CVF Conference on Computer Vision
  and Pattern Recognition}, 2022.

\bibitem[Oord et~al.(2018)Oord, Li, and Vinyals]{oord2018representation}
A.~v.~d. Oord, Y.~Li, and O.~Vinyals.
\newblock Representation learning with contrastive predictive coding.
\newblock \emph{arXiv preprint arXiv:1807.03748}, 2018.

\bibitem[Osman et~al.(2020)Osman, Bolkart, and Black]{osman2020star}
A.~A. Osman, T.~Bolkart, and M.~J. Black.
\newblock Star: Sparse trained articulated human body regressor.
\newblock In \emph{European Conference on Computer Vision}, pages 598--613.
  Springer, 2020.

\bibitem[Pang et~al.(2018)Pang, Li, Song, Song, Xiang, and
  Hospedales]{pang2018deep}
K.~Pang, D.~Li, J.~Song, Y.-Z. Song, T.~Xiang, and T.~M. Hospedales.
\newblock Deep factorised inverse-sketching.
\newblock In \emph{Proceedings of the European Conference on Computer Vision
  (ECCV)}, pages 36--52, 2018.

\bibitem[Papandreou et~al.(2018)Papandreou, Zhu, Chen, Gidaris, Tompson, and
  Murphy]{papandreou2018personlab}
G.~Papandreou, T.~Zhu, L.-C. Chen, S.~Gidaris, J.~Tompson, and K.~Murphy.
\newblock Personlab: Person pose estimation and instance segmentation with a
  bottom-up, part-based, geometric embedding model.
\newblock In \emph{Proceedings of the European conference on computer vision
  (ECCV)}, pages 269--286, 2018.

\bibitem[Pavlakos et~al.(2019{\natexlab{a}})Pavlakos, Choutas, Ghorbani,
  Bolkart, Osman, Tzionas, and Black]{pavlakos2019expressive}
G.~Pavlakos, V.~Choutas, N.~Ghorbani, T.~Bolkart, A.~A. Osman, D.~Tzionas, and
  M.~J. Black.
\newblock Expressive body capture: 3d hands, face, and body from a single
  image.
\newblock In \emph{Proceedings of the IEEE/CVF conference on computer vision
  and pattern recognition}, pages 10975--10985, 2019{\natexlab{a}}.

\bibitem[Pavlakos et~al.(2019{\natexlab{b}})Pavlakos, Choutas, Ghorbani,
  Bolkart, Osman, Tzionas, and Black]{SMPLX2019}
G.~Pavlakos, V.~Choutas, N.~Ghorbani, T.~Bolkart, A.~A.~A. Osman, D.~Tzionas,
  and M.~J. Black.
\newblock Expressive body capture: {3D} hands, face, and body from a single
  image.
\newblock In \emph{Proceedings IEEE Conf. on Computer Vision and Pattern
  Recognition (CVPR)}, pages 10975--10985, 2019{\natexlab{b}}.

\bibitem[Ren et~al.(2019)Ren, Yu, Zhang, Li, Liu, and Li]{ren2019structureflow}
Y.~Ren, X.~Yu, R.~Zhang, T.~H. Li, S.~Liu, and G.~Li.
\newblock Structureflow: Image inpainting via structure-aware appearance flow.
\newblock In \emph{Proceedings of the IEEE/CVF International Conference on
  Computer Vision}, pages 181--190, 2019.

\bibitem[Rhodin et~al.(2018)Rhodin, Salzmann, and Fua]{rhodin2018unsupervised}
H.~Rhodin, M.~Salzmann, and P.~Fua.
\newblock Unsupervised geometry-aware representation for 3d human pose
  estimation.
\newblock In \emph{Proceedings of the European Conference on Computer Vision
  (ECCV)}, pages 750--767, 2018.

\bibitem[Rhodin et~al.(2019)Rhodin, Constantin, Katircioglu, Salzmann, and
  Fua]{Rhodin_2019_CVPR}
H.~Rhodin, V.~Constantin, I.~Katircioglu, M.~Salzmann, and P.~Fua.
\newblock Neural scene decomposition for multi-person motion capture.
\newblock In \emph{Proceedings of the IEEE/CVF Conference on Computer Vision
  and Pattern Recognition (CVPR)}, June 2019.

\bibitem[Riza Alp~G\"uler(2018)]{Guler2018DensePose}
I.~K. Riza Alp~G\"uler, Natalia~Neverova.
\newblock Densepose: Dense human pose estimation in the wild.
\newblock In \emph{The IEEE Conference on Computer Vision and Pattern
  Recognition (CVPR)}, 2018.

\bibitem[Ronneberger et~al.(2015)Ronneberger, Fischer, and
  Brox]{ronneberger2015u}
O.~Ronneberger, P.~Fischer, and T.~Brox.
\newblock U-net: Convolutional networks for biomedical image segmentation.
\newblock In \emph{International Conference on Medical image computing and
  computer-assisted intervention}, pages 234--241. Springer, 2015.

\bibitem[Sangkloy et~al.(2016)Sangkloy, Burnell, Ham, and
  Hays]{sangkloy2016the}
P.~Sangkloy, N.~Burnell, C.~Ham, and J.~Hays.
\newblock The sketchy database: Learning to retrieve badly drawn bunnies.
\newblock \emph{ACM Transactions on Graphics (proceedings of SIGGRAPH)}, 2016.

\bibitem[Sasaki et~al.(2018)Sasaki, Iizuka, Simo-Serra, and
  Ishikawa]{SasakiCGI2018learning}
K.~Sasaki, S.~Iizuka, E.~Simo-Serra, and H.~Ishikawa.
\newblock {Learning to Restore Deteriorated Line Drawing}.
\newblock \emph{The Visual Computer (Proc. of Computer Graphics International
  2018)}, 34\penalty0 (6-8):\penalty0 1077--1085, 2018.

\bibitem[Savarese et~al.(2021)Savarese, Kim, Maire, Shakhnarovich, and
  McAllester]{savarese2021information}
P.~Savarese, S.~S. Kim, M.~Maire, G.~Shakhnarovich, and D.~McAllester.
\newblock Information-theoretic segmentation by inpainting error maximization.
\newblock In \emph{Proceedings of the IEEE/CVF Conference on Computer Vision
  and Pattern Recognition}, pages 4029--4039, 2021.

\bibitem[Schmidtke et~al.(2021)Schmidtke, Vlontzos, Ellershaw, Lukens, Arichi,
  and Kainz]{schmidtke2021unsupervised}
L.~Schmidtke, A.~Vlontzos, S.~Ellershaw, A.~Lukens, T.~Arichi, and B.~Kainz.
\newblock Unsupervised human pose estimation through transforming shape
  templates.
\newblock In \emph{Proceedings of the IEEE/CVF Conference on Computer Vision
  and Pattern Recognition}, pages 2484--2494, 2021.

\bibitem[Siarohin et~al.(2018)Siarohin, Sangineto, Lathuiliere, and
  Sebe]{siarohin2018deformable}
A.~Siarohin, E.~Sangineto, S.~Lathuiliere, and N.~Sebe.
\newblock Deformable gans for pose-based human image generation.
\newblock In \emph{Proceedings of the IEEE Conference on Computer Vision and
  Pattern Recognition}, pages 3408--3416, 2018.

\bibitem[Siarohin et~al.(2019{\natexlab{a}})Siarohin, Lathuili{\`e}re,
  Tulyakov, Ricci, and Sebe]{siarohin2019animating}
A.~Siarohin, S.~Lathuili{\`e}re, S.~Tulyakov, E.~Ricci, and N.~Sebe.
\newblock Animating arbitrary objects via deep motion transfer.
\newblock In \emph{Proceedings of the IEEE Conference on Computer Vision and
  Pattern Recognition}, pages 2377--2386, 2019{\natexlab{a}}.

\bibitem[Siarohin et~al.(2019{\natexlab{b}})Siarohin, Lathuilière, Tulyakov,
  Ricci, and Sebe]{Siarohin_2019_NeurIPS}
A.~Siarohin, S.~Lathuilière, S.~Tulyakov, E.~Ricci, and N.~Sebe.
\newblock First order motion model for image animation.
\newblock In \emph{Conference on Neural Information Processing Systems
  (NeurIPS)}, December 2019{\natexlab{b}}.

\bibitem[Siarohin et~al.(2021)Siarohin, Roy, Lathuili{\`e}re, Tulyakov, Ricci,
  and Sebe]{siarohin2021motion}
A.~Siarohin, S.~Roy, S.~Lathuili{\`e}re, S.~Tulyakov, E.~Ricci, and N.~Sebe.
\newblock Motion-supervised co-part segmentation.
\newblock In \emph{2020 25th International Conference on Pattern Recognition
  (ICPR)}, pages 9650--9657. IEEE, 2021.

\bibitem[Song et~al.(2018)Song, Yang, Shen, Wang, Huang, and Kuo]{song2018spg}
Y.~Song, C.~Yang, Y.~Shen, P.~Wang, Q.~Huang, and C.-C.~J. Kuo.
\newblock Spg-net: Segmentation prediction and guidance network for image
  inpainting.
\newblock \emph{arXiv preprint arXiv:1805.03356}, 2018.

\bibitem[Su et~al.(2021)Su, Yu, Zollhoefer, and Rhodin]{su2021anerf}
S.-Y. Su, F.~Yu, M.~Zollhoefer, and H.~Rhodin.
\newblock A-ne{RF}: Articulated neural radiance fields for learning human
  shape, appearance, and pose.
\newblock In A.~Beygelzimer, Y.~Dauphin, P.~Liang, and J.~W. Vaughan, editors,
  \emph{Advances in Neural Information Processing Systems}, 2021.
\newblock URL \url{https://openreview.net/forum?id=lwwEh0OM61b}.

\bibitem[Sun et~al.(2022)Sun, Ryou, Goldshmid, Weissbourd, Dabiri, Anderson,
  Kennedy, Yue, and Perona]{sun2022self}
J.~J. Sun, S.~Ryou, R.~H. Goldshmid, B.~Weissbourd, J.~O. Dabiri, D.~J.
  Anderson, A.~Kennedy, Y.~Yue, and P.~Perona.
\newblock Self-supervised keypoint discovery in behavioral videos.
\newblock In \emph{Proceedings of the IEEE/CVF Conference on Computer Vision
  and Pattern Recognition}, pages 2171--2180, 2022.

\bibitem[Suwajanakorn et~al.(2018)Suwajanakorn, Snavely, Tompson, and
  Norouzi]{suwajanakorn2018discovery}
S.~Suwajanakorn, N.~Snavely, J.~J. Tompson, and M.~Norouzi.
\newblock Discovery of latent 3d keypoints via end-to-end geometric reasoning.
\newblock In \emph{Advances in neural information processing systems}, pages
  2059--2070, 2018.

\bibitem[Thewlis et~al.(2017)Thewlis, Bilen, and
  Vedaldi]{thewlis2017unsupervised}
J.~Thewlis, H.~Bilen, and A.~Vedaldi.
\newblock Unsupervised learning of object landmarks by factorized spatial
  embeddings.
\newblock In \emph{Proceedings of the IEEE international conference on computer
  vision}, pages 5916--5925, 2017.

\bibitem[Wah et~al.(2011)Wah, Branson, Welinder, Perona, and
  Belongie]{WahCUB_200_2011}
C.~Wah, S.~Branson, P.~Welinder, P.~Perona, and S.~Belongie.
\newblock {The Caltech-UCSD Birds-200-2011 Dataset}.
\newblock Technical Report CNS-TR-2011-001, California Institute of Technology,
  2011.

\bibitem[Wang et~al.(2015)Wang, Kang, and Li]{wang2015sketch}
F.~Wang, L.~Kang, and Y.~Li.
\newblock Sketch-based 3d shape retrieval using convolutional neural networks.
\newblock In \emph{Proceedings of the IEEE conference on computer vision and
  pattern recognition}, pages 1875--1883, 2015.

\bibitem[Wang et~al.(2019)Wang, Ge, Lipton, and Xing]{wang2019learning}
H.~Wang, S.~Ge, Z.~Lipton, and E.~P. Xing.
\newblock Learning robust global representations by penalizing local predictive
  power.
\newblock In \emph{Advances in Neural Information Processing Systems}, pages
  10506--10518, 2019.

\bibitem[Wang et~al.(2020)Wang, Wang, Huang, Shi, Cai, Zhu, and
  Yin]{wang2020image}
J.~Wang, C.~Wang, Q.~Huang, Y.~Shi, J.-F. Cai, Q.~Zhu, and B.~Yin.
\newblock \emph{Image Inpainting Based on Multi-Frequency Probabilistic
  Inference Model}, page 1–9.
\newblock Association for Computing Machinery, New York, NY, USA, 2020.
\newblock ISBN 9781450379885.
\newblock URL \url{https://doi.org/10.1145/3394171.3413891}.

\bibitem[Wu et~al.(2019)Wu, Cao, Li, Qian, and Loy]{wu2019transgaga}
W.~Wu, K.~Cao, C.~Li, C.~Qian, and C.~C. Loy.
\newblock Transgaga: Geometry-aware unsupervised image-to-image translation.
\newblock In \emph{Proceedings of the IEEE Conference on Computer Vision and
  Pattern Recognition}, pages 8012--8021, 2019.

\bibitem[Wu et~al.(2018)Wu, Xiong, Stella, and Lin]{wu2018unsupervised}
Z.~Wu, Y.~Xiong, X.~Y. Stella, and D.~Lin.
\newblock Unsupervised feature learning via non-parametric instance
  discrimination.
\newblock In \emph{Proceedings of the IEEE Conference on Computer Vision and
  Pattern Recognition}, 2018.

\bibitem[Xiao et~al.(2018)Xiao, Wu, and Wei]{xiao2018simple}
B.~Xiao, H.~Wu, and Y.~Wei.
\newblock Simple baselines for human pose estimation and tracking.
\newblock In \emph{European Conference on Computer Vision (ECCV)}, 2018.

\bibitem[Xiong et~al.(2019)Xiong, Yu, Lin, Yang, Lu, Barnes, and
  Luo]{xiong2019foreground}
W.~Xiong, J.~Yu, Z.~Lin, J.~Yang, X.~Lu, C.~Barnes, and J.~Luo.
\newblock Foreground-aware image inpainting.
\newblock In \emph{Proceedings of the IEEE/CVF Conference on Computer Vision
  and Pattern Recognition}, pages 5840--5848, 2019.

\bibitem[Xu et~al.(2020{\natexlab{a}})Xu, Bazavan, Zanfir, Freeman, Sukthankar,
  and Sminchisescu]{Xu2020ghum}
H.~Xu, E.~G. Bazavan, A.~Zanfir, B.~Freeman, R.~Sukthankar, and
  C.~Sminchisescu.
\newblock Ghum \& ghuml: Generative 3d human shape and articulated pose models.
\newblock In \emph{IEEE/CVF Conference on Computer Vision and Pattern
  Recognition (Oral)}, pages 6184--6193, 2020{\natexlab{a}}.
\newblock URL
  \url{https://openaccess.thecvf.com/content_CVPR_2020/html/Xu_GHUM__GHUML_Generative_3D_Human_Shape_and_Articulated_Pose_CVPR_2020_paper.html}.

\bibitem[Xu et~al.(2020{\natexlab{b}})Xu, Su, Wang, Hao, and Gao]{xu2020anedge}
H.~Xu, X.~Su, M.~Wang, X.~Hao, and G.~Gao.
\newblock An edge information and mask shrinking based image inpainting
  approach.
\newblock In \emph{2020 IEEE International Conference on Multimedia and Expo
  (ICME)}, pages 1--6, Los Alamitos, CA, USA, jul 2020{\natexlab{b}}. IEEE
  Computer Society.
\newblock \doi{10.1109/ICME46284.2020.9102892}.
\newblock URL
  \url{https://doi.ieeecomputersociety.org/10.1109/ICME46284.2020.9102892}.

\bibitem[Xu(2022)]{Xu2022DeepLF}
P.~Xu.
\newblock Deep learning for free-hand sketch: A survey.
\newblock \emph{IEEE transactions on pattern analysis and machine
  intelligence}, PP, 2022.

\bibitem[Xu et~al.(2018)Xu, Huang, Yuan, Pang, Song, Xiang, Hospedales, Ma, and
  Guo]{xu2018sketchmate}
P.~Xu, Y.~Huang, T.~Yuan, K.~Pang, Y.-Z. Song, T.~Xiang, T.~M. Hospedales,
  Z.~Ma, and J.~Guo.
\newblock Sketchmate: Deep hashing for million-scale human sketch retrieval.
\newblock In \emph{Proceedings of the IEEE conference on computer vision and
  pattern recognition}, pages 8090--8098, 2018.

\bibitem[Xu et~al.(2021)Xu, Joshi, and Bresson]{xu2021multigraph}
P.~Xu, C.~K. Joshi, and X.~Bresson.
\newblock Multigraph transformer for free-hand sketch recognition.
\newblock \emph{IEEE Transactions on Neural Networks and Learning Systems},
  2021.

\bibitem[Xu et~al.(2020{\natexlab{c}})Xu, Yang, Liu, Dai, and
  Zhou]{xu2020unsupervised}
Y.~Xu, C.~Yang, Z.~Liu, B.~Dai, and B.~Zhou.
\newblock Unsupervised landmark learning from unpaired data.
\newblock \emph{arXiv preprint arXiv:2007.01053}, 2020{\natexlab{c}}.

\bibitem[Yang and Guo(2020)]{yang2020generative}
Y.~Yang and X.~Guo.
\newblock Generative landmark guided face inpainting.
\newblock In \emph{Chinese Conference on Pattern Recognition and Computer
  Vision (PRCV)}, pages 14--26. Springer, 2020.

\bibitem[Yang et~al.(2019)Yang, Loquercio, Scaramuzza, and
  Soatto]{yang2019unsupervised}
Y.~Yang, A.~Loquercio, D.~Scaramuzza, and S.~Soatto.
\newblock Unsupervised moving object detection via contextual information
  separation.
\newblock In \emph{Proceedings of the IEEE/CVF Conference on Computer Vision
  and Pattern Recognition}, pages 879--888, 2019.

\bibitem[Yang et~al.(2021)Yang, Lai, and Soatto]{yang2021dystab}
Y.~Yang, B.~Lai, and S.~Soatto.
\newblock Dystab: Unsupervised object segmentation via dynamic-static
  bootstrapping.
\newblock In \emph{Proceedings of the IEEE/CVF Conference on Computer Vision
  and Pattern Recognition}, pages 2826--2836, 2021.

\bibitem[Yi et~al.(2019)Yi, Liu, Lai, and Rosin]{YiLLR19APDrawingGAN}
R.~Yi, Y.-J. Liu, Y.-K. Lai, and P.~L. Rosin.
\newblock {APDrawingGAN}: Generating artistic portrait drawings from face
  photos with hierarchical gans.
\newblock In \emph{{IEEE} Conference on Computer Vision and Pattern Recognition
  (CVPR '19)}, pages 10743--10752, 2019.

\bibitem[Yu et~al.(2015)Yu, Yang, Song, Xiang, and
  Hospedales]{Yu2015SketchaNetTB}
Q.~Yu, Y.~Yang, Y.-Z. Song, T.~Xiang, and T.~M. Hospedales.
\newblock Sketch-a-net that beats humans.
\newblock In \emph{BMVC}, 2015.

\bibitem[Yu et~al.(2016)Yu, Liu, Song, Xiang, Hospedales, and
  Loy]{qian2016sketch}
Q.~Yu, F.~Liu, Y.-Z. Song, T.~Xiang, T.~M. Hospedales, and C.~C. Loy.
\newblock Sketch me that shoe.
\newblock In \emph{IEEE Conference on Computer Vision and Pattern Recognition
  (CVPR)}, 2016.

\bibitem[Yu et~al.(2017)Yu, Guo, Xu, Dong, Su, Zhao, Li, Dai, and
  Liu]{yu2017bodyfusion}
T.~Yu, K.~Guo, F.~Xu, Y.~Dong, Z.~Su, J.~Zhao, J.~Li, Q.~Dai, and Y.~Liu.
\newblock Bodyfusion: Real-time capture of human motion and surface geometry
  using a single depth camera.
\newblock In \emph{2017 IEEE International Conference on Computer Vision
  (ICCV)}, pages 910--919, 2017.
\newblock \doi{10.1109/ICCV.2017.104}.

\bibitem[Yu et~al.(2021)Yu, Zhan, WU, Pan, Cui, Lu, Ma, Xie, and
  Miao]{yu2021diverse}
Y.~Yu, F.~Zhan, R.~WU, J.~Pan, K.~Cui, S.~Lu, F.~Ma, X.~Xie, and C.~Miao.
\newblock \emph{Diverse Image Inpainting with Bidirectional and Autoregressive
  Transformers}, page 69–78.
\newblock Association for Computing Machinery, New York, NY, USA, 2021.
\newblock ISBN 9781450386517.
\newblock URL \url{https://doi.org/10.1145/3474085.3475436}.

\bibitem[Zhang et~al.(2021)Zhang, Shi, Wang, Wu, Li, Lv, and
  Mumtaz]{zhang2021face}
X.~Zhang, C.~Shi, X.~Wang, X.~Wu, X.~Li, J.~Lv, and I.~Mumtaz.
\newblock Face inpainting based on gan by facial prediction and fusion as
  guidance information.
\newblock \emph{Applied Soft Computing}, 111:\penalty0 107626, 2021.
\newblock ISSN 1568-4946.
\newblock \doi{https://doi.org/10.1016/j.asoc.2021.107626}.
\newblock URL
  \url{https://www.sciencedirect.com/science/article/pii/S1568494621005470}.

\bibitem[Zhang et~al.(2018)Zhang, Guo, Jin, Luo, He, and
  Lee]{zhang2018unsupervised}
Y.~Zhang, Y.~Guo, Y.~Jin, Y.~Luo, Z.~He, and H.~Lee.
\newblock Unsupervised discovery of object landmarks as structural
  representations.
\newblock In \emph{Proceedings of the IEEE Conference on Computer Vision and
  Pattern Recognition}, pages 2694--2703, 2018.

\bibitem[Zhang et~al.(2022)Zhang, Liang, Zou, Li, Sun, and Wang]{zhang2022self}
Y.~Zhang, Q.~Liang, K.~Zou, Z.~Li, W.~Sun, and Y.~Wang.
\newblock Self-supervised part segmentation via motion imitation.
\newblock \emph{Image and Vision Computing}, 120:\penalty0 104393, 2022.
\newblock ISSN 0262-8856.
\newblock \doi{https://doi.org/10.1016/j.imavis.2022.104393}.
\newblock URL
  \url{https://www.sciencedirect.com/science/article/pii/S0262885622000221}.

\bibitem[Zhao et~al.(2021)Zhao, Liu, Xu, Chen, Luo, Jin, Zhu, Liu, Zhao, and
  Gao]{zhao2021prior}
Z.~Zhao, W.~Liu, Y.~Xu, X.~Chen, W.~Luo, L.~Jin, B.~Zhu, T.~Liu, B.~Zhao, and
  S.~Gao.
\newblock Prior based human completion.
\newblock In \emph{Proceedings of the IEEE/CVF International Conference on
  Computer Vision}, pages 7947--7957, 2021.

\bibitem[Zheng et~al.(2019)Zheng, Cham, and Cai]{zheng2019pluralistic}
C.~Zheng, T.-J. Cham, and J.~Cai.
\newblock Pluralistic image completion.
\newblock In \emph{Proceedings of the IEEE/CVF Conference on Computer Vision
  and Pattern Recognition}, pages 1438--1447, 2019.

\bibitem[Zhu et~al.(2017)Zhu, Park, Isola, and Efros]{zhu2017unpaired}
J.-Y. Zhu, T.~Park, P.~Isola, and A.~A. Efros.
\newblock Unpaired image-to-image translation using cycle-consistent
  adversarial networks.
\newblock In \emph{Proceedings of the IEEE international conference on computer
  vision}, pages 2223--2232, 2017.

\bibitem[Zhu et~al.(2014)Zhu, Liang, Wei, and Sun]{zhu2014saliency}
W.~Zhu, S.~Liang, Y.~Wei, and J.~Sun.
\newblock Saliency optimization from robust background detection.
\newblock In \emph{2014 IEEE Conference on Computer Vision and Pattern
  Recognition}, pages 2814--2821, 2014.
\newblock \doi{10.1109/CVPR.2014.360}.

\end{thebibliography}

\clearpage

\clearpage
\appendix

\section{More Results}  \label{supp:more_results}
We show supplemental videos in \url{https://xingzhehe.github.io/autolink/}.

To verify the robustness and generality, we show 105 images for each dataset we used in the main paper with keypoints and visualized graph representation in Figure~\ref{fig:gen_afhq_k32}-\ref{fig:taichi_k32} in the supplemental materials.

\section{Edge-Map Ablation Tests} \label{supp:ablation_test}

Table~\ref{tab:hyper} shows the results on the different numbers of keypoints and edge thickness. While a larger number of keypoints gives better details and thus higher accuracy, the performance is robust to the thickness.
Table~\ref{tab:mask_strategy} shows the results on different masking ratios and mask patch sizes. A too small masking ratio significantly decreases the performance since the structure can be directly extracted from the masked image with a low masking ratio. The thicknesses used in the main paper are those marked bold in Table~\ref{tab:hyper}. \looseness=-1

\begin{table}[h]
\centering
\resizebox{0.98\linewidth}{!}{%
\begin{tabular}{|c|c|c|c|c|c|c|c|c|c|c|c|c|c|c|c|c|c|c|c|c|}
\hline
 & \multicolumn{5}{c|}{CelebA-wild $\downarrow$} & \multicolumn{5}{c|}{Human3.6m $\downarrow$} & \multicolumn{5}{c|}{DeepFashion $\uparrow$} & \multicolumn{5}{c|}{Taichi $\downarrow$} \\ \hline
$\sigma^2$ &  K=2 & K=4  & K=8  & K=16 & K=32 &  K=2 &  K=4 & K=8  &K=16 &  K=32 &  K=2 &  K=4 &  K=8 & K=16 & K=32 & K=2 & K=4 & K=8 & K=16& K=32    \\ \cline{2-21}
    1.0e-5 & 60.9 & 52.5 & 42.6 & 34.9 & 31.9 & 5.67 & 5.13 & 3.37 & 2.90 & \textbf{2.81} & 17.8 & 48.1 & 58.1 & 62.6 & 63.8 & 667 & 657 & 622 & 592 & 458 \\
    2.5e-5 & 50.5 & 38.8 & 29.8 & 10.7 & 6.11 & 5.64 & 5.15 & 3.50 & 3.02 & 2.87 & 29.5 & 48.9 & 57.1 & 65.2 & 66.2 & 665 & 637 & 611 & 506 & 351 \\
    5.0e-5 & 49.4 & 8.06 & 5.41 & 4.88 & 4.65 & 6.10 & 5.03 & 3.76 & \textbf{2.76} & 2.91 & 22.5 & 50.4 & 58.7 & \textbf{65.8} & 66.6 & 654 & 550 & 338 & 338 & 287 \\ 
    7.5e-5 & 58.3 & 7.71 & 5.62 & 4.92 & 4.65 & 5.49 & 5.08 & \textbf{3.19} & 2.89 & 3.00 & 43.5 & 51.1 & \textbf{59.3} & 65.7 & \textbf{69.8} & 650 & 516 & 383 & 301 & 297 \\
    1.0e-4 & 54.5 & 7.44 & 5.71 & 4.93 & 4.64 & 5.56 & 5.09 & 3.25 & 2.96 & 2.96 & 44.5 & 49.0 & 58.1 & 64.1 & 67.6 & 647 & 512 & 385 & 329 & 280 \\
    2.5e-4 & 11.3 & 6.68 & 5.57 & 5.05 & 4.42 & 5.49 & 5.03 & 3.36 & 3.01 & 3.32 & 42.9 & 48.7 & 56.8 & 64.9 & 67.6 & 624 & 526 & 391 & 321 & 284 \\
    5.0e-4 & 11.4 & 6.56 & 5.77 & 5.01 & 4.43 & 5.49 & 5.10 & 3.45 & 2.94 & 3.16 & 42.0 & 47.5 & 58.0 & 65.4 & 67.2 & 612 & 531 & 381 & 307 & \textbf{275} \\
    7.5e-4 & \textbf{10.6} & \textbf{6.11} & 5.81 & 4.86 & \textbf{4.39} & 5.57 & 5.09 & 3.84 & 3.31 & 3.06 & 35.6 & 49.6 & 58.2 & 65.7 & 66.4 & 604 & 479 & 363 & 296 & 289 \\
    1.0e-3 & 13.3 & 6.84 & 5.70 & \textbf{4.43} & 4.50 & 5.49 & 5.04 & 3.42 & 3.47 & 3.18 & 40.4 & 51.0 & 57.1 & 64.6 & 68.8 & 608 & 462 & 342 & 306 & 275 \\
    2.5e-3 & 15.8 & 6.70 & \textbf{5.24} & 4.72 & 4.49 & 5.48 & 5.04 & 3.40 & 3.37 & 3.01 & 43.6 & \textbf{51.6} & 55.8 & 61.2 & 66.7 & 609 & \textbf{442} & 326 & \textbf{289} & 286 \\
    5.0e-3 & 12.6 & 6.29 & 5.53 & 4.69 & 4.48 & 5.52 & 5.03 & 3.59 & 3.36 & 3.12 & \textbf{44.7} & 50.3 & 56.7 & 62.3 & 68.3 & 598 & 510 & 333 & 323 & 302 \\
    7.5e-3 & 11.6 & 6.16 & 5.61 & 4.76 & 4.41 & 5.54 & 5.06 & 3.50 & 3.77 & 2.99 & 39.4 & 48.6 & 57.7 & 62.6 & 65.5 & 604 & 514 & \textbf{325} & 328 & 340 \\
    1.0e-2 & 11.2 & 6.22 & 6.66 & 4.89 & 4.47 & \textbf{5.45} & \textbf{5.02} & 3.39 & 3.05 & 2.94 & 44.6 & 50.3 & 56.5 & 60.9 & 65.1 & \textbf{593} & 515 & 327 & 327 & 341 \\ \hline
\end{tabular}
}
\caption{\textbf{Ablation Tests on the numbers of keypoints and edge thickness}. We remove the \% sign in metrics for simplicity. The best one for each number of keypoints is marked in bold.} 
\label{tab:hyper}
\end{table}

\begin{table}[h]
\centering
\resizebox{0.98\linewidth}{!}{%
\begin{tabular}{|c|c|c|c|c|c|c|c|c|c|c|c|c|c|c|c|c|c|c|c|c|}
\hline
 & \multicolumn{5}{c|}{CelebA-wild $\downarrow$} & \multicolumn{5}{c|}{Human3.6m $\downarrow$} & \multicolumn{5}{c|}{DeepFashion $\uparrow$} & \multicolumn{5}{c|}{Taichi $\downarrow$} \\ \hline
ratio & 4x4  & 8x8  & 16x16& 32x32& 64x64& 4x4  & 8x8  & 16x16& 32x32& 64x64& 4x4  & 8x8  & 16x16& 32x32& 64x64& 4x4 & 8x8 &16x16&32x32& 64x64    \\ \cline{2-21}
10\%  & 17.9 & 16.4 & 49.2 & 49.3 & 48.3 & 3.22 & 3.25 & 3.55 & 5.28 & 6.39 & 58.3 & 39.7 & 40.5 & 52.1 & 51.0 & 640 & 638 & 630 & 659 & 642 \\
20\%  & 19.7 & 29.4 & 44.6 & 43.3 & 39.8 & 3.06 & 3.52 & 3.21 & 4.29 & 5.99 & 62.1 & 55.5 & 43.4 & 39.6 & 49.3 & 522 & 605 & 614 & 638 & 613 \\
30\%  & 6.56 & 8.46 & 45.4 & 41.3 & 46.8 & 3.08 & 3.08 & 3.44 & 3.80 & 6.24 & 62.8 & 58.9 & 62.8 & 42.4 & 41.8 & 509 & 570 & 627 & 642 & 627 \\ 
40\%  & 8.31 & 6.71 & 7.58 & 6.31 & 6.81 & 3.15 & 2.91 & 3.38 & 3.40 & 4.19 & 62.5 & 59.2 & 63.4 & 41.2 & 40.0 & 500 & 428 & 634 & 640 & 651 \\
50\%  & 8.06 & 6.39 & 6.23 & 6.58 & 5.99 & 3.73 & 2.99 & 2.94 & 3.72 & 4.24 & 61.3 & 65.4 & 64.3 & 41.1 & 38.9 & 444 & 452 & 481 & 640 & 667 \\
60\%  & 7.69 & 6.22 & 5.52 & 6.98 & 5.56 & 3.11 & 3.13 & 2.95 & 3.35 & 4.87 & 62.8 & 62.0 & 61.7 & 41.4 & 39.9 & 483 & 396 & 418 & 654 & 643 \\
70\%  & 7.03 & 6.38 & 5.44 & 5.11 & 4.43 & 2.95 & 3.09 & 2.87 & 3.28 & 4.85 & 60.7 & 63.2 & 63.1 & 59.7 & 41.5 & 447 & 362 & 376 & 523 & 656 \\
80\%  & 6.95 & 6.68 & 5.24 & 4.73 & 4.65 & 3.63 & 3.47 & 2.76 & 2.97 & 4.08 & 58.3 & 61.7 & 65.8 & 61.6 & 39.6 & 501 & 371 & 316 & 347 & 642 \\
90\%  & 7.24 & 7.15 & 5.77 & 5.62 & 4.14 & 2.99 & 3.18 & 2.95 & 3.64 & 3.75 & 60.9 & 63.1 & 66.4 & 62.7 & 40.2 & 626 & 388 & 330 & 346 & 526 \\ \hline
\end{tabular}
}
\caption{\textbf{Ablation Tests on masking ratio and patch size}. We remove the \% sign in metrics for simplicity. The best one for each number of keypoints is marked in bold.} 
\label{tab:mask_strategy}
\end{table}

\section{Applications}
In this section, we briefly describe how we created the two applications, conditional Generative Adversarial Networks and pose transfer networks, based on the learned graph representation, as shown in the teaser. Note that, although we apply the graph representation to videos for pose transfer, it is only learned from the collections of single images.

\subsection{Conditional Generative Adversarial Network}
The conditional GAN is a simplified StyleGAN2 \cite{karras2020analyzing}, where the spatial noise injection is removed and the starting tensor is replaced by the feature map generated from the edge map. For simplicity, we do not use EqualLinear \cite{karras2019style} or Path Regularization \cite{karras2019style}.

Formally speaking, given an edge map $\mS\in\R^{H\times W}$, where $H, W$ are the spatial size, we use bicubic interpolation to downsample it to $32\times 32$ and feed it into a two-layer convolution network to generate a feature map $\mF\in\R^{8\times 8\times 512}$. In parallel, we generate the embedding vector $\vw\in\R^{256}$ from a noise vector $\vz\in\R^{256}$ by a three linear-layer MLP. The feature map $\mF$ is fed into a residual convolution-based generator where the kernel weights are modulated by the embedding vector $\vw$ \cite{karras2020analyzing}. The final image $\vx\in\R^{H\times W\times 3}$ is generated by a single convolution layer.

We denote $\cG$ as the generator and $\cD$ as the discriminator. We use the non-saturating loss \cite{goodfellow2014generative},
\begin{equation}
    \cL_\text{GAN}(\cG)=\mathbb E_{\vz\sim\cN}\log(\exp(-\cD(\cG(\vz)))+1)
\label{eq:gen_loss}
\end{equation}
for the generator, and logistic loss,
\begin{equation}
        \cL_\text{GAN}(\cD)=\mathbb E_{\vz\sim\cN}\log(\exp(\cD(\cG(\vz)))+1)+\mathbb E_{\vx\sim p_\text{data}}\log(\exp(-\cD(\vx))+1)
    \label{eq:dis_loss}
\end{equation}
for the discriminator, with gradient penalty~\cite{mescheder2018training} applied only on real data,
\begin{equation}
    \cL_\text{gp}(\cD)=\mathbb E_{\vx\sim p_\text{data}}\nabla\cD(\vx).
\label{eq:gradient penalty}
\end{equation}

\subsection{Pose Transfer Network}
We train the pose transfer network on videos. We randomly sample two frames $\mI_1, \mI_2\in\R^{H\times W\times 3}$, where $H,W$ are the spatial size. We use keypoint detector on the frame $\mI_1$ and generate the edge map $\mS_1\in\R^{H\times W}$. The edge map is fed into a UNet \cite{ronneberger2015u} to reconstruct $\mI_1$. The smallest feature map in the UNet is concatenated by a feature map of the appearance information, which is generated by frame $\mI_2$ to provide the appearance information. The loss is a combination of Mean Squared Error, VGG perceptual loss \cite{johnson2016perceptual} and the GAN loss in Eq~\ref{eq:gen_loss}, \ref{eq:dis_loss}, and \ref{eq:gradient penalty}.

\section{Network Architecture} \label{supp:archi}
\begin{figure}[t]
\begin{center}
  \includegraphics[width=0.98\textwidth]{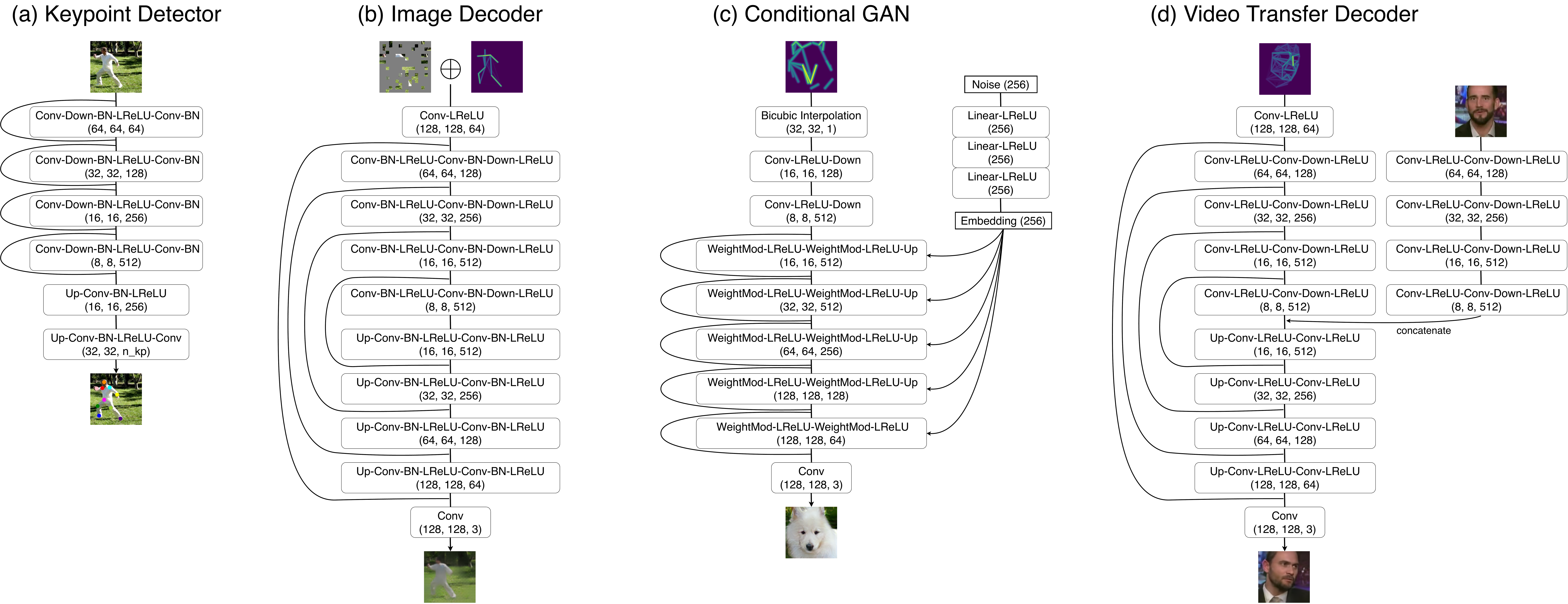}
\end{center}
    \caption{\textbf{Network architectures. From left to right: detector (encoder), decoder, conditional GAN, conditional autoencoder.}
    The shortcuts in (a) and (c) are addition while the shortcuts in (b) and (d) are concatenation.
    }
\label{fig:net_archi}
\end{figure}

Figure~\ref{fig:net_archi} shows the architectures we used in the main paper. 
The keypoint detector is a ResNet with upsampling \cite{xiao2018simple}, which is a simple baseline used in human pose estimation.
The decoder and the pose transfer network are UNets \cite{ronneberger2015u}.
The conditional GAN is a simplified StyleGAN2 \cite{karras2020analyzing}, where the spatial noise injection is removed and the starting tensor is replaced by the feature map generated from the edge map.
In Figure~\ref{fig:net_archi}, we denote Conv for 3x3 convolution \cite{lecun1989backpropagation}, BN for Batch Normalization \cite{ioffe2015batch}, LReLU for Leaky ReLU \cite{maas2013rectifier}, Up for 2x bilinear upsampling, and Down for 2x bilinear downsampling. 

\section{Edge Map Visualization}
For visualization purposes, we scale the edge weights by dividing the maximum value to obtain visible edges. For DeepFashion, before dividing the maximum value, we further add 0.01 to the edge weights larger than 0.0001. Although the models are trained on different thicknesses, we draw them in the same thickness $\sigma^2=5\times 10^{-4}$ for pleasing visualization.

\begin{figure*}[t]
\begin{center}
   \includegraphics[width=1\linewidth]{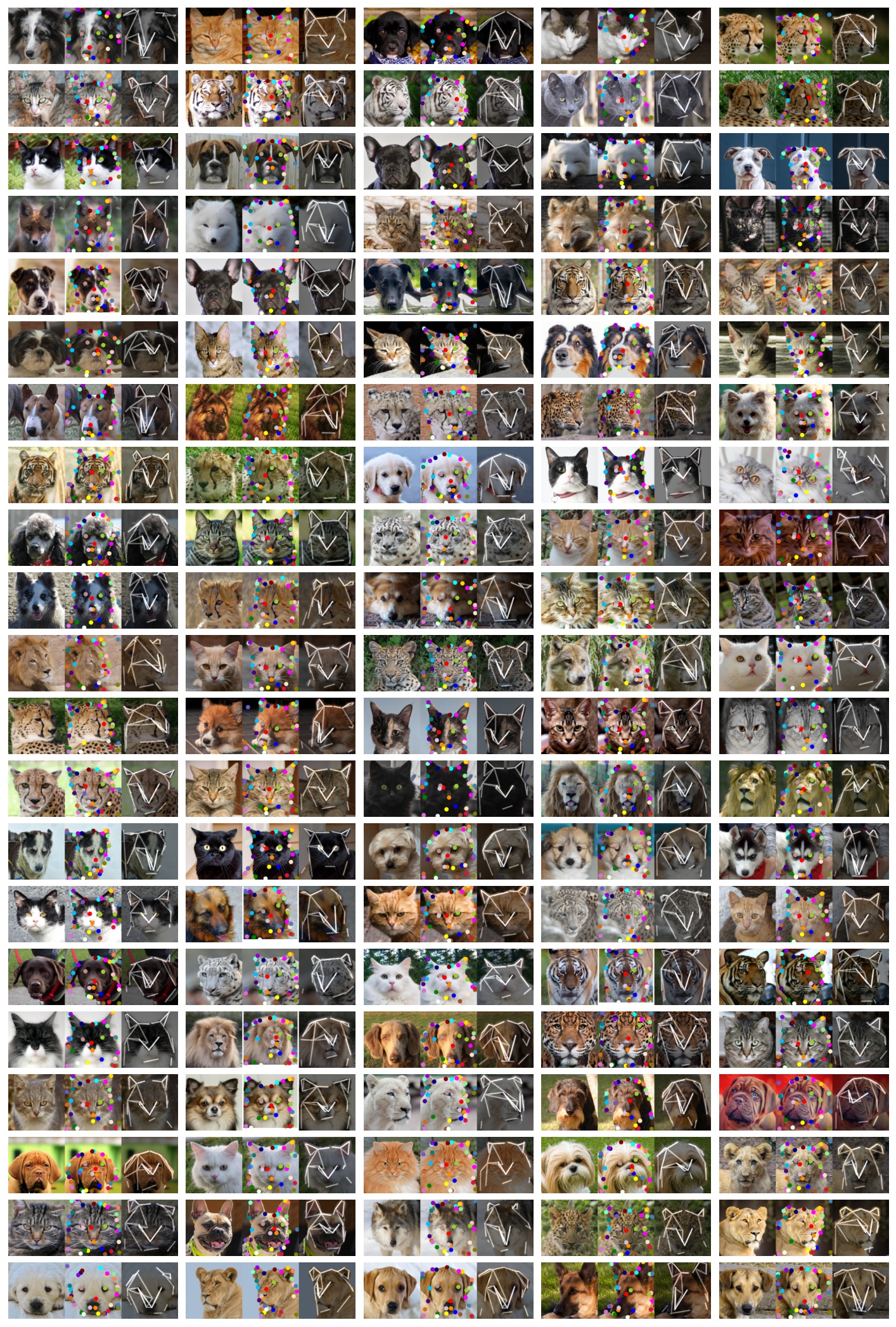}
\end{center}
   \caption{\textbf{105 samples from AFHQ (32 keypoints),} with the image-points-edge pairs overlaid.}
   \label{fig:gen_afhq_k32}
\end{figure*}

\begin{figure*}[t]
\begin{center}
   \includegraphics[width=1\linewidth]{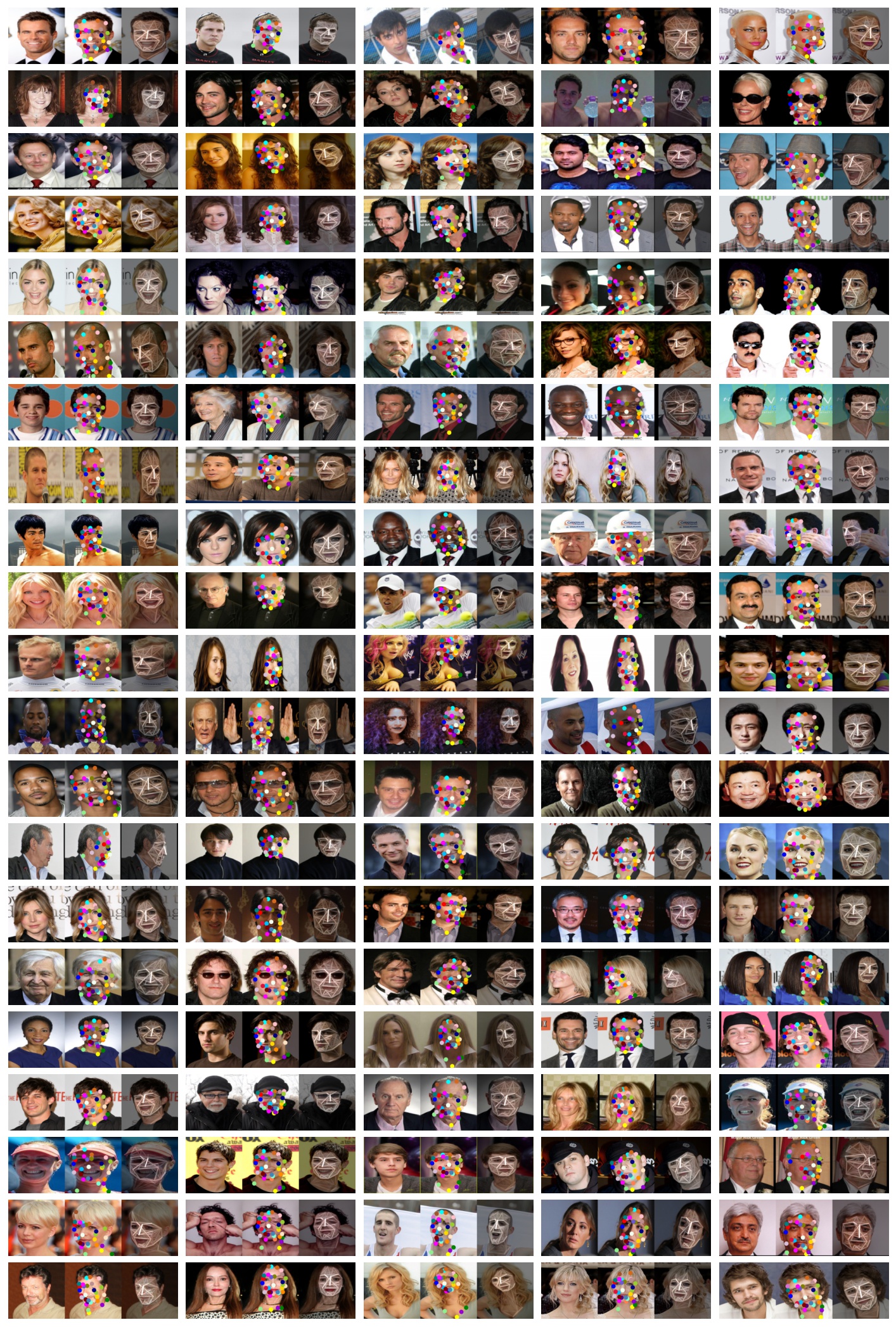}
\end{center}
   \caption{\textbf{105 samples from CelebA-in-The-Wild (32 keypoints),} with the image-points-edge pairs overlaid.}
   \label{fig:gen_face_k32}
\end{figure*}

\begin{figure*}[t]
\begin{center}
   \includegraphics[width=1\linewidth]{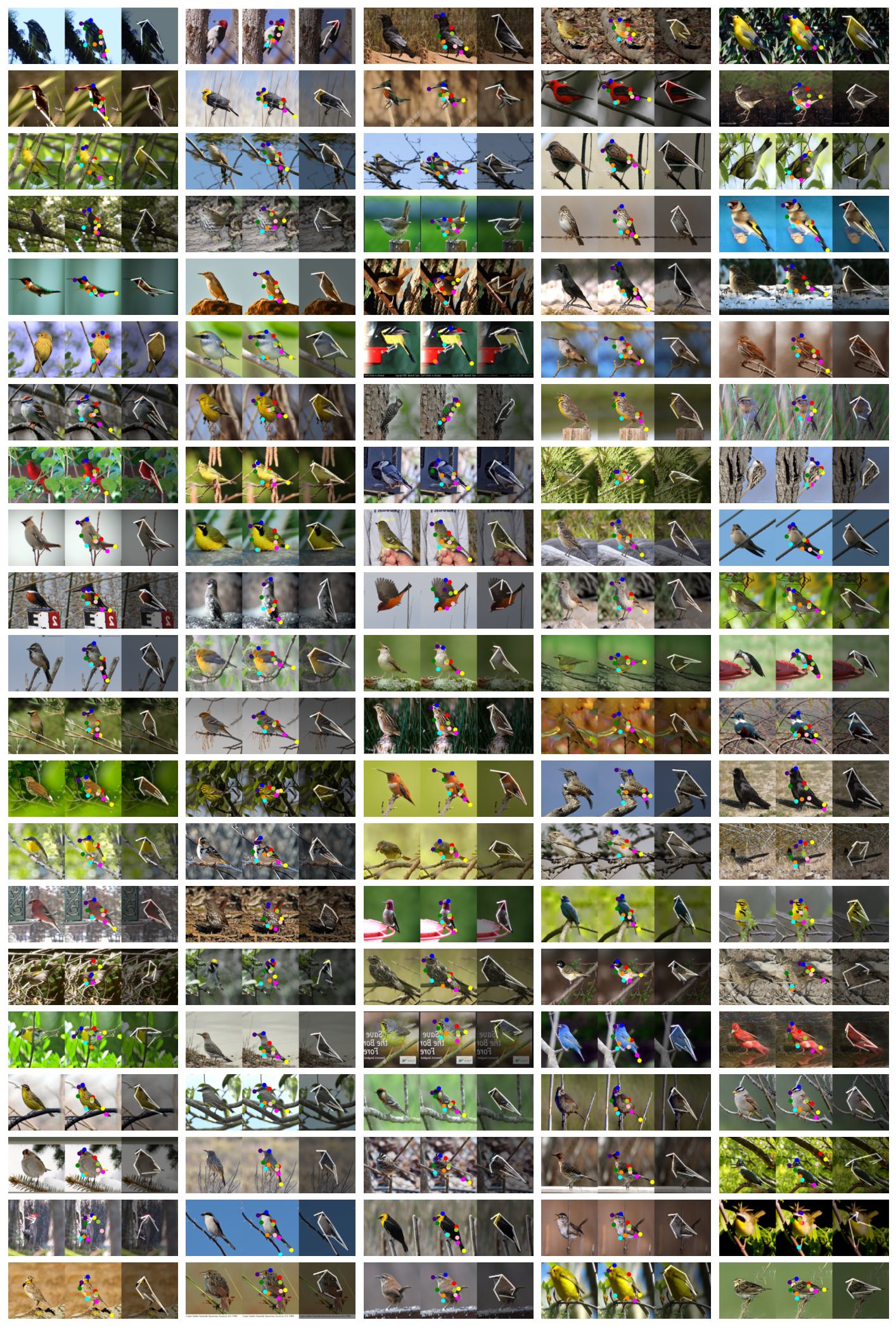}
\end{center}
   \caption{\textbf{105 samples from CUB-aligned (10 keypoints),} with the image-points-edge pairs overlaid.}
\end{figure*}

\begin{figure*}[t]
\begin{center}
   \includegraphics[width=1\linewidth]{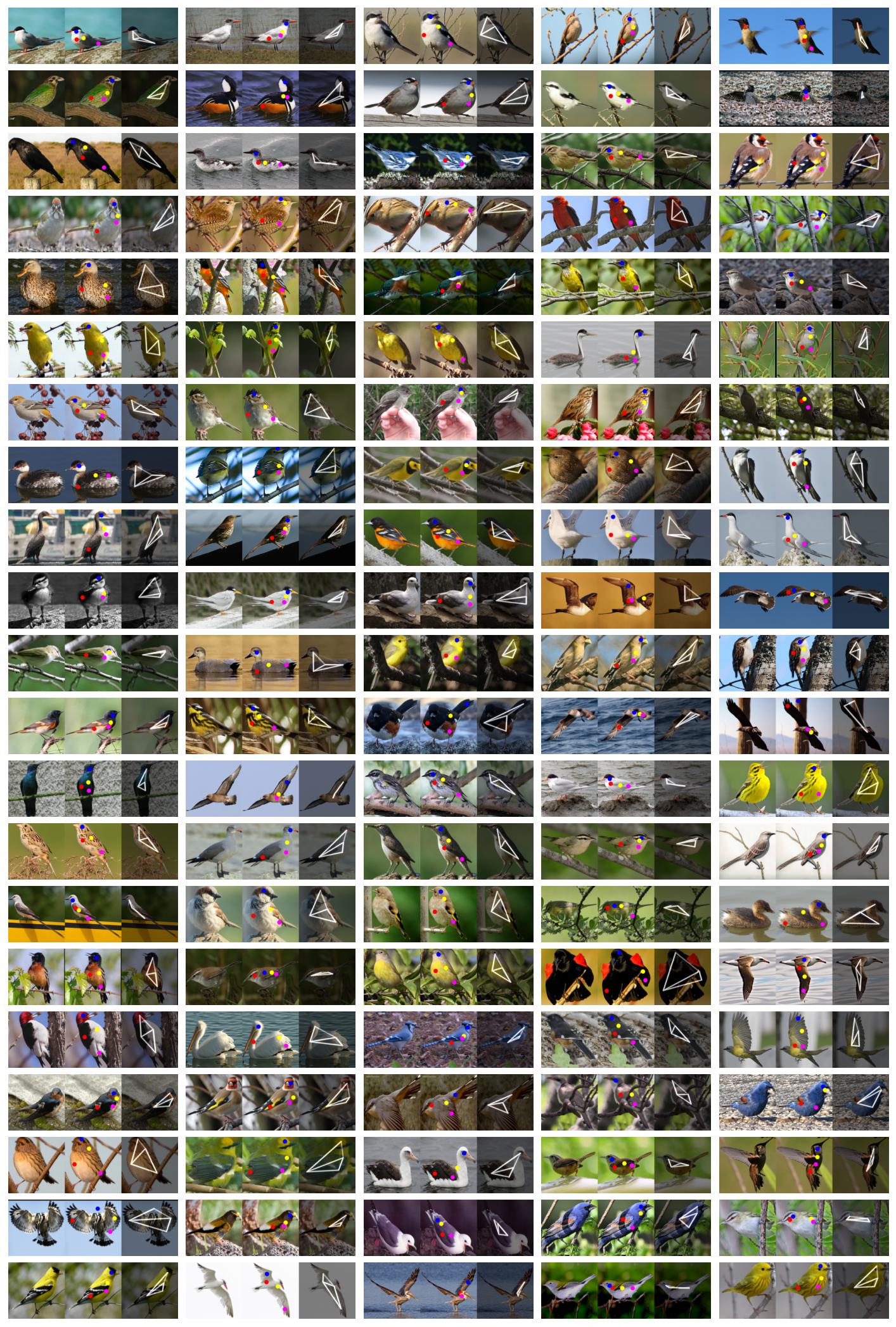}
\end{center}
   \caption{\textbf{105 samples from CUB (4 keypoints),} with the image-points-edge pairs overlaid.}
\end{figure*}

\begin{figure*}[t]
\begin{center}
   \includegraphics[width=1\linewidth]{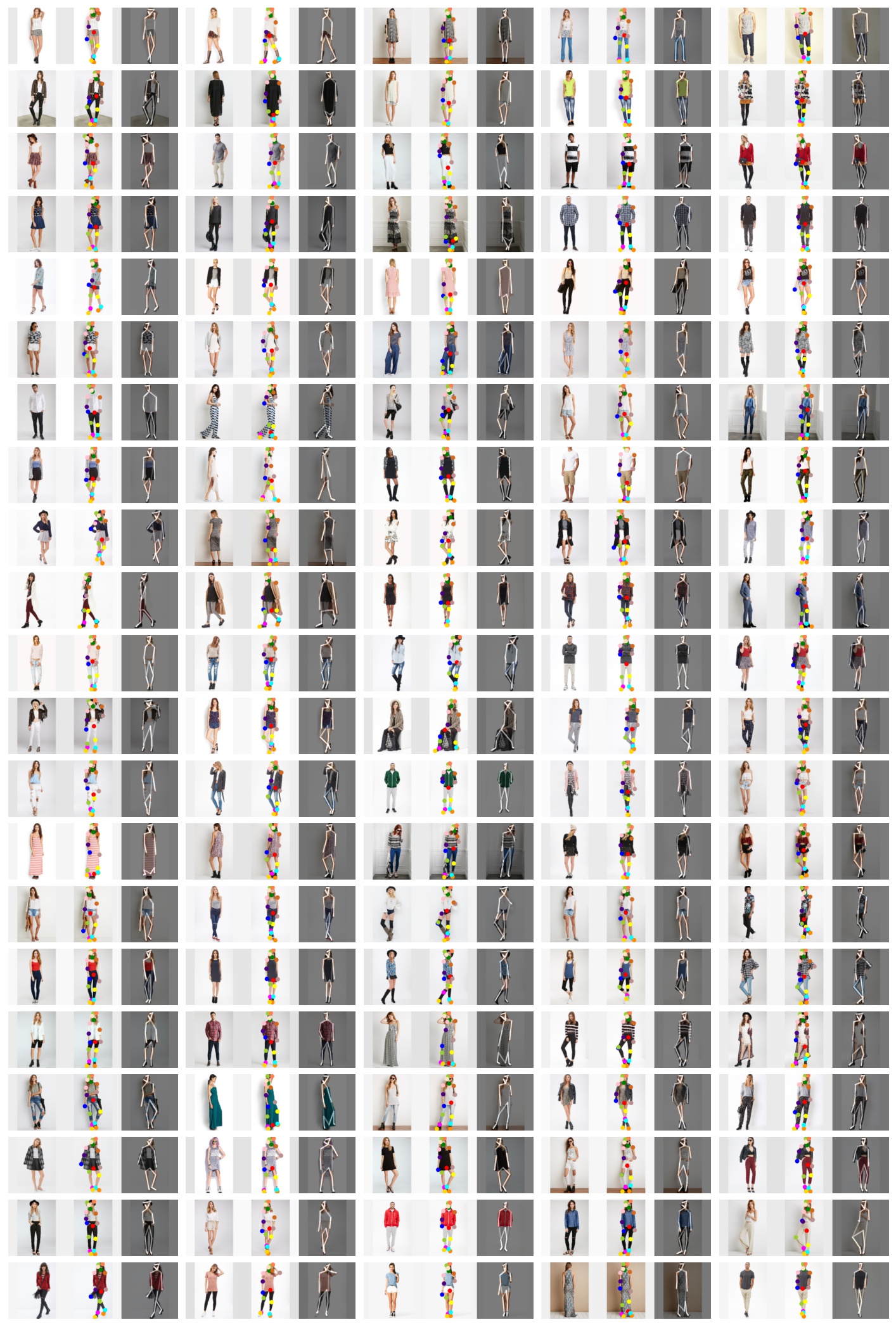}
\end{center}
   \caption{\textbf{105 samples from DeepFashion (16 keypoints),} with the image-points-edge pairs overlaid.}
   \label{fig:gen_deepfashion_k32}
\end{figure*}

\begin{figure*}[t]
\begin{center}
   \includegraphics[width=1\linewidth]{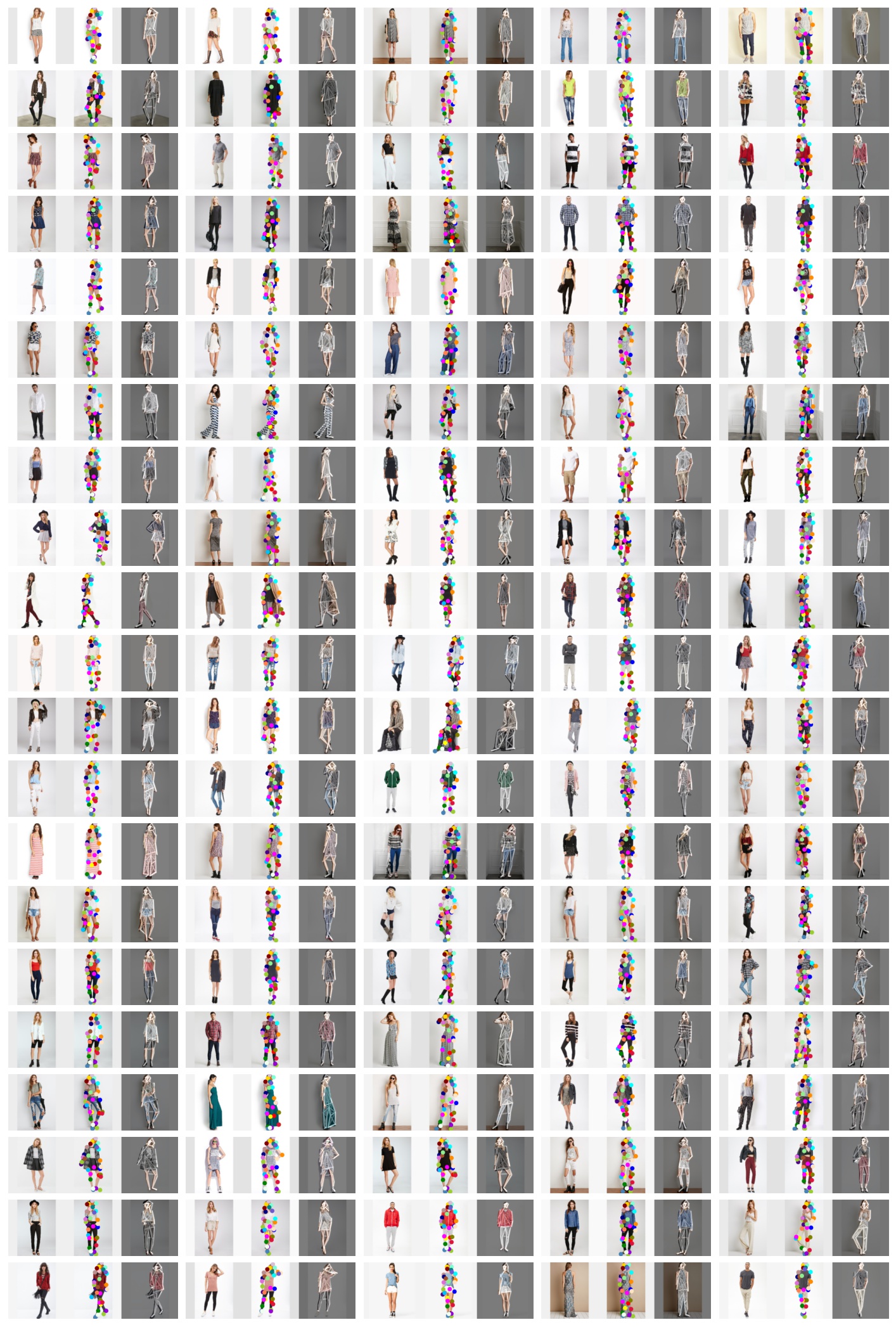}
\end{center}
   \caption{\textbf{105 samples from DeepFashion (32 keypoints),} with the image-points-edge pairs overlaid.}
\end{figure*}

\begin{figure*}[t]
\begin{center}
   \includegraphics[width=1\linewidth]{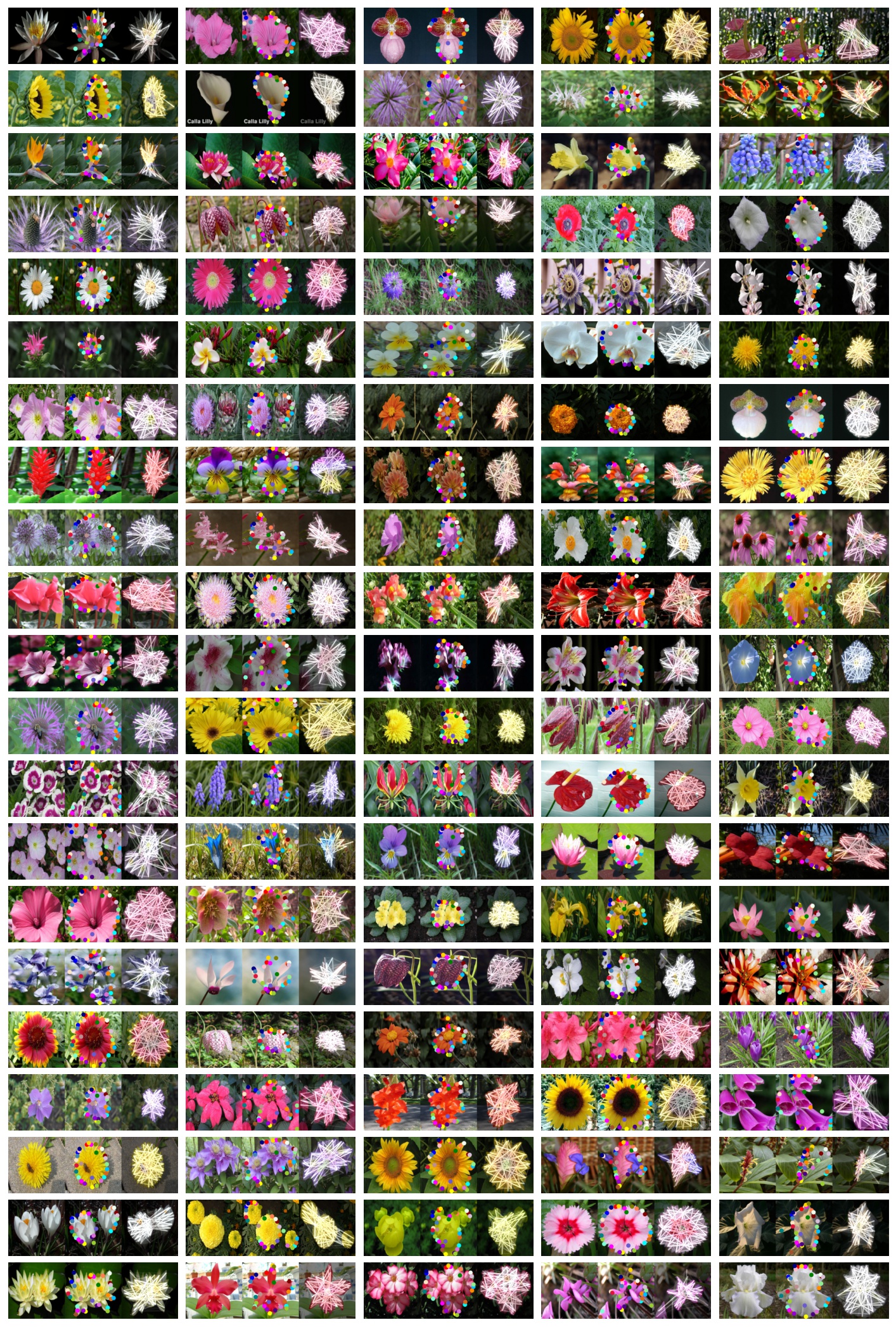}
\end{center}
   \caption{\textbf{105 samples from Flower (32 keypoints),} with the image-points-edge pairs overlaid.}
\end{figure*}

\begin{figure*}[t]
\begin{center}
   \includegraphics[width=1\linewidth]{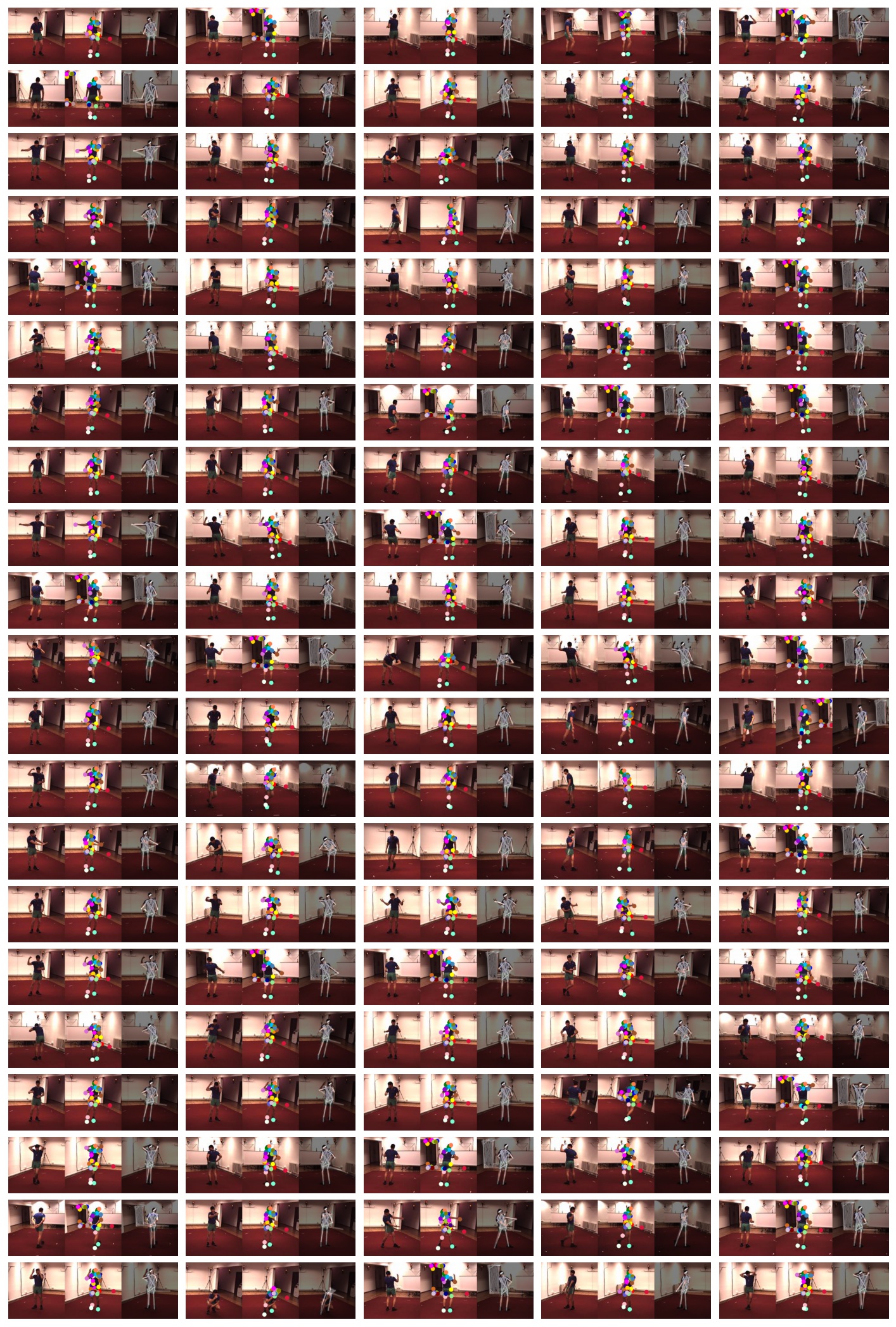}
\end{center}
   \caption{\textbf{105 samples from Human3.6m (32 keypoints),} with the image-points-edge pairs overlaid.}
\end{figure*}

\begin{figure*}[t]
\begin{center}
   \includegraphics[width=1\linewidth]{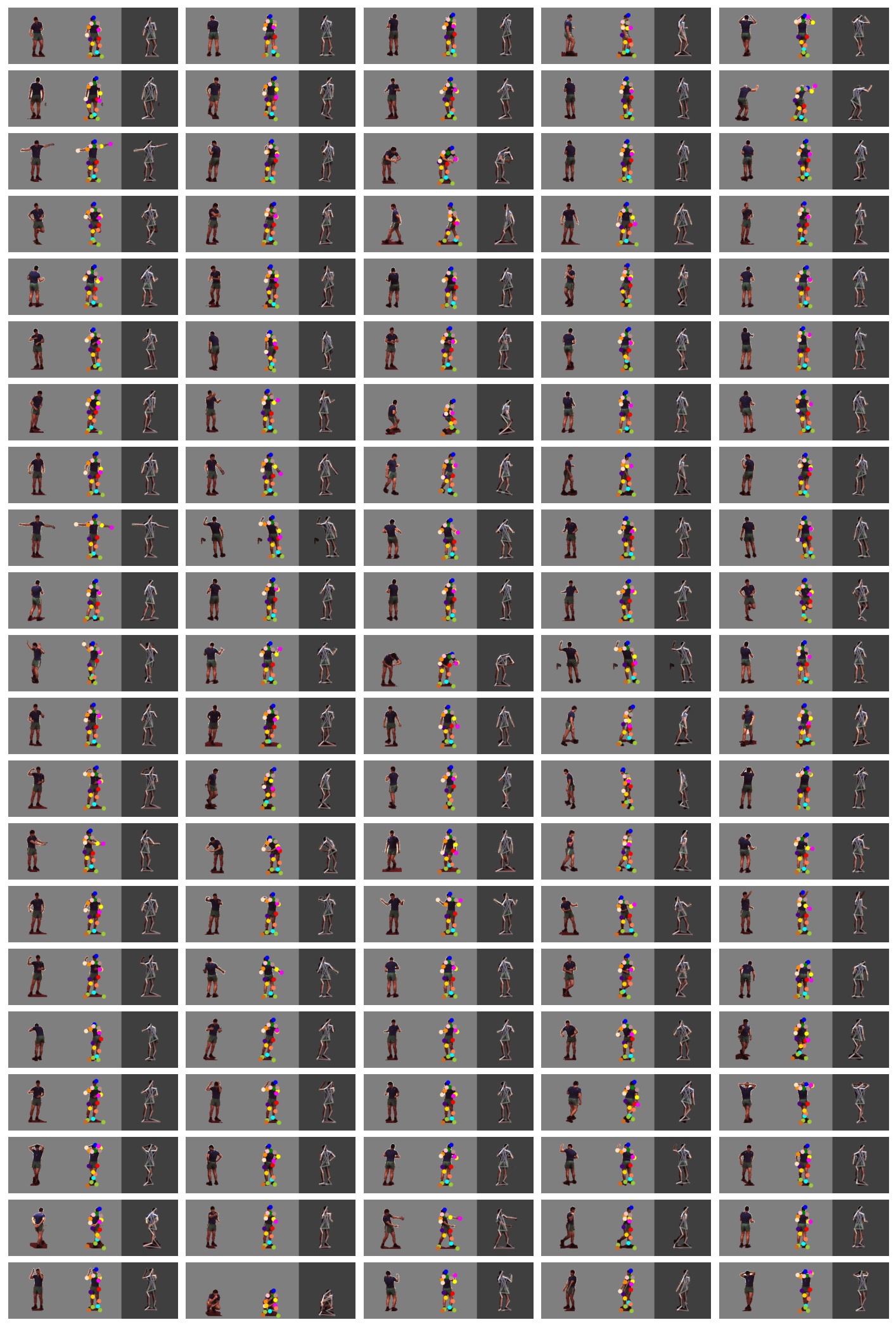}
\end{center}
   \caption{\textbf{105 samples from Human3.6m without background (16 keypoints),} with the image-points-edge pairs overlaid.}
\end{figure*}

\begin{figure*}[t]
\begin{center}
   \includegraphics[width=1\linewidth]{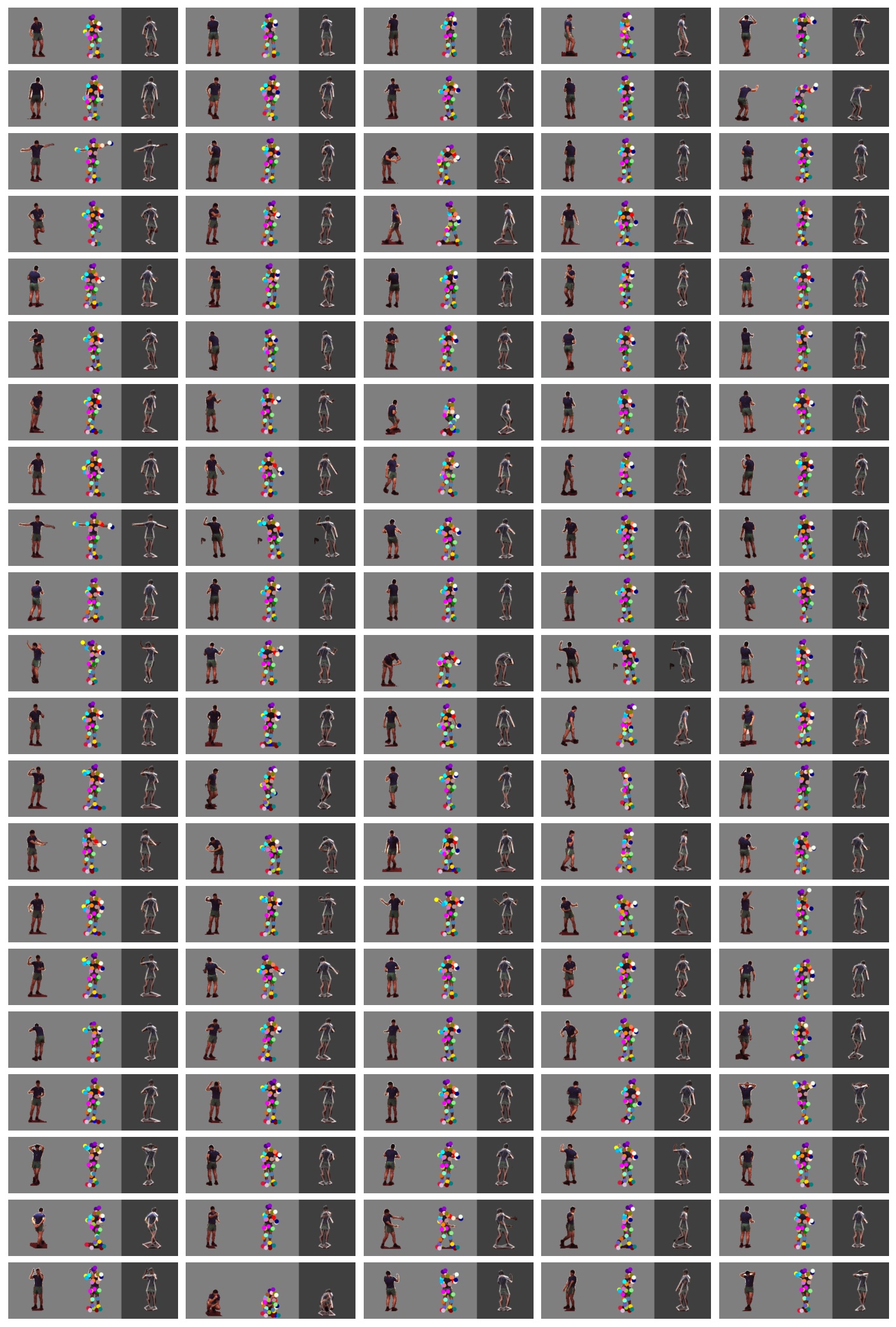}
\end{center}
   \caption{\textbf{105 samples from Human3.6m without background (32 keypoints),} with the image-points-edge pairs overlaid.}
\end{figure*}

\begin{figure*}[t]
\begin{center}
   \includegraphics[width=1\linewidth]{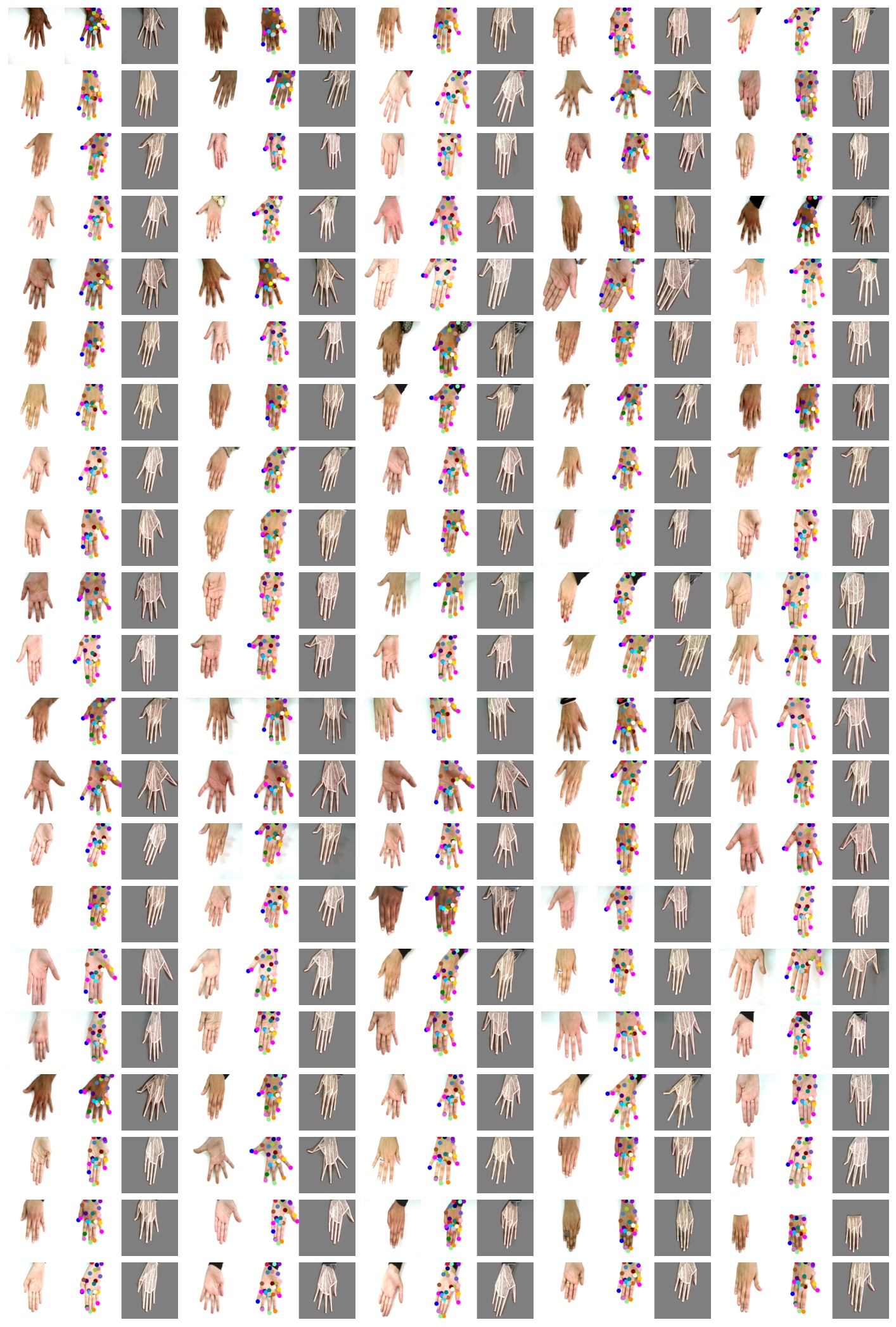}
\end{center}
   \caption{\textbf{105 samples from 11k Hands without background (32 keypoints),} with the image-points-edge pairs overlaid.}
\end{figure*}

\begin{figure*}[t]
\begin{center}
   \includegraphics[width=1\linewidth]{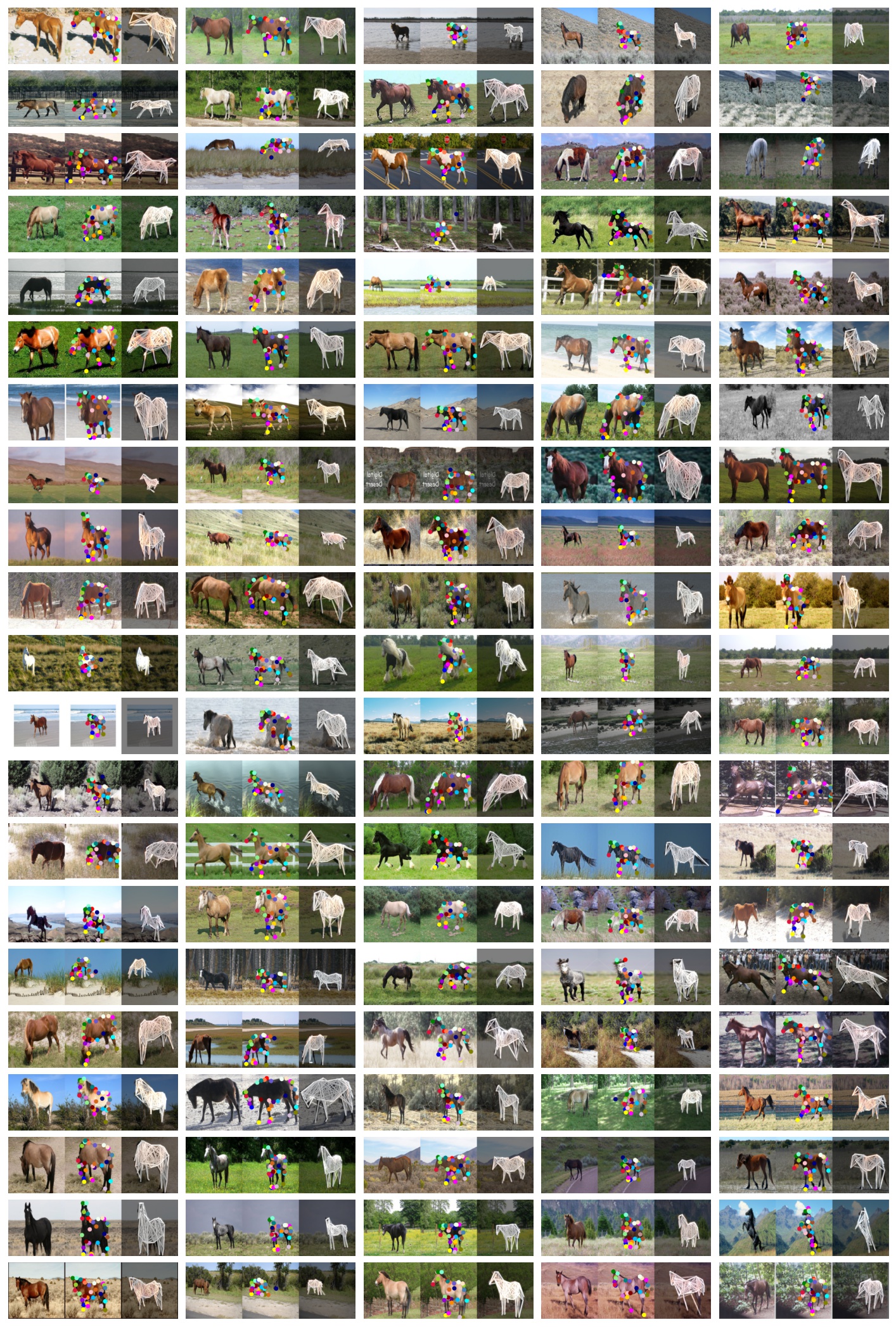}
\end{center}
   \caption{\textbf{105 samples from Horses (32 keypoints),} with the image-points-edge pairs overlaid.}
\end{figure*}

\begin{figure*}[t]
\begin{center}
   \includegraphics[width=1\linewidth]{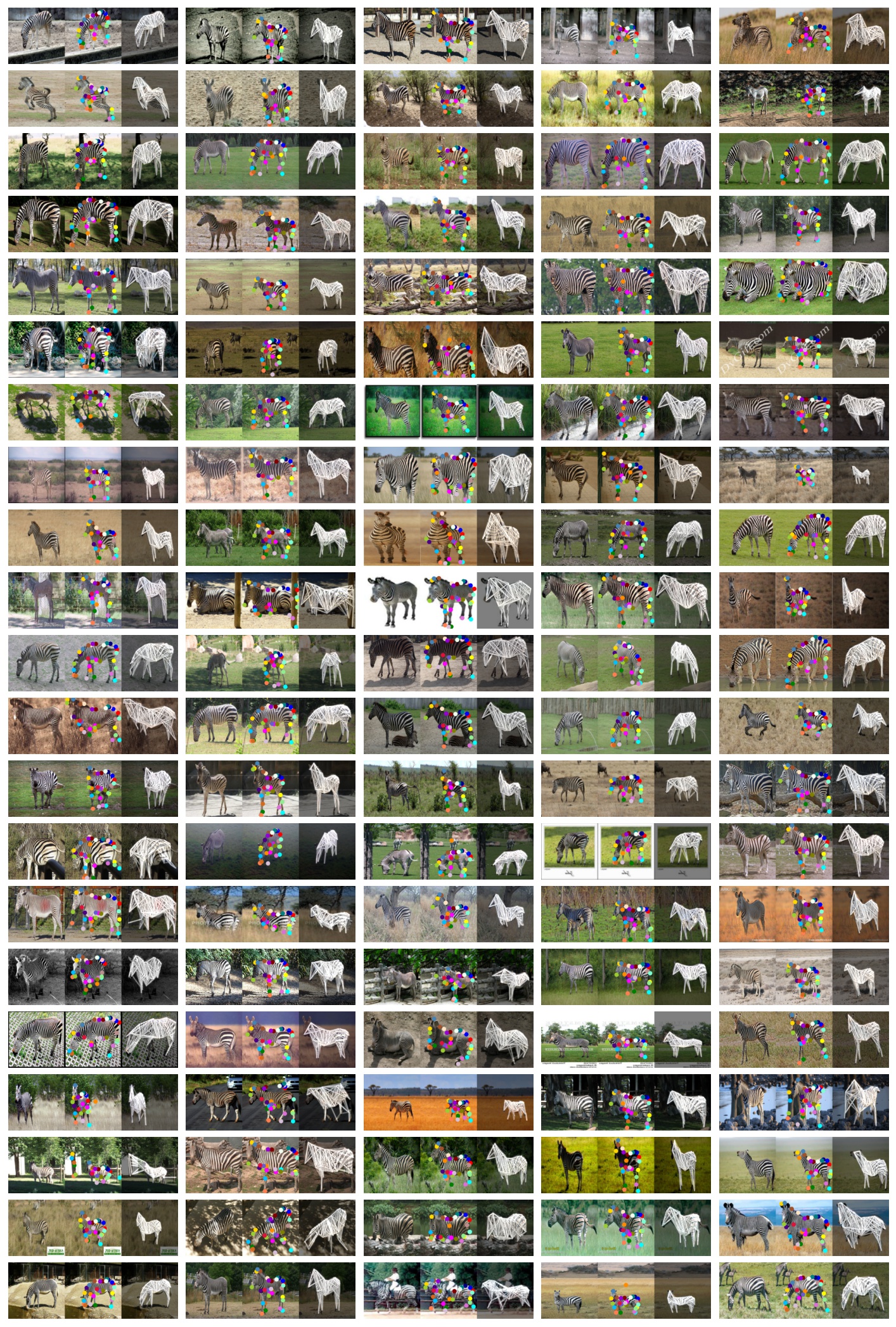}
\end{center}
   \caption{\textbf{105 samples from Zebra (32 keypoints),} with the image-points-edge pairs overlaid.}
\end{figure*}

\begin{figure*}[t]
\begin{center}
   \includegraphics[width=1\linewidth]{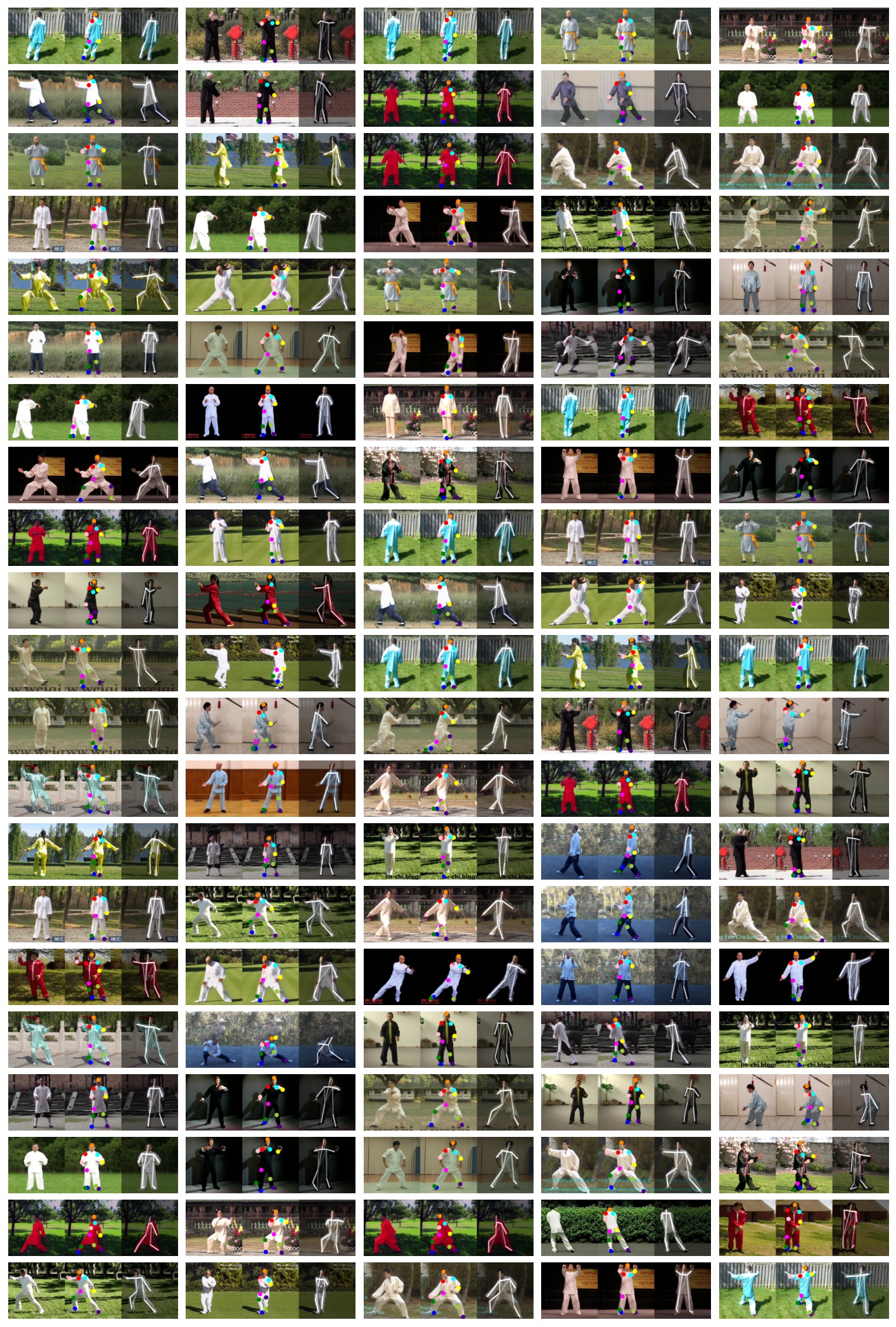}
\end{center}
   \caption{\textbf{105 samples from Taichi (10 keypoints),} with the image-points-edge pairs overlaid.}
\end{figure*}

\begin{figure*}[t]
\begin{center}
   \includegraphics[width=1\linewidth]{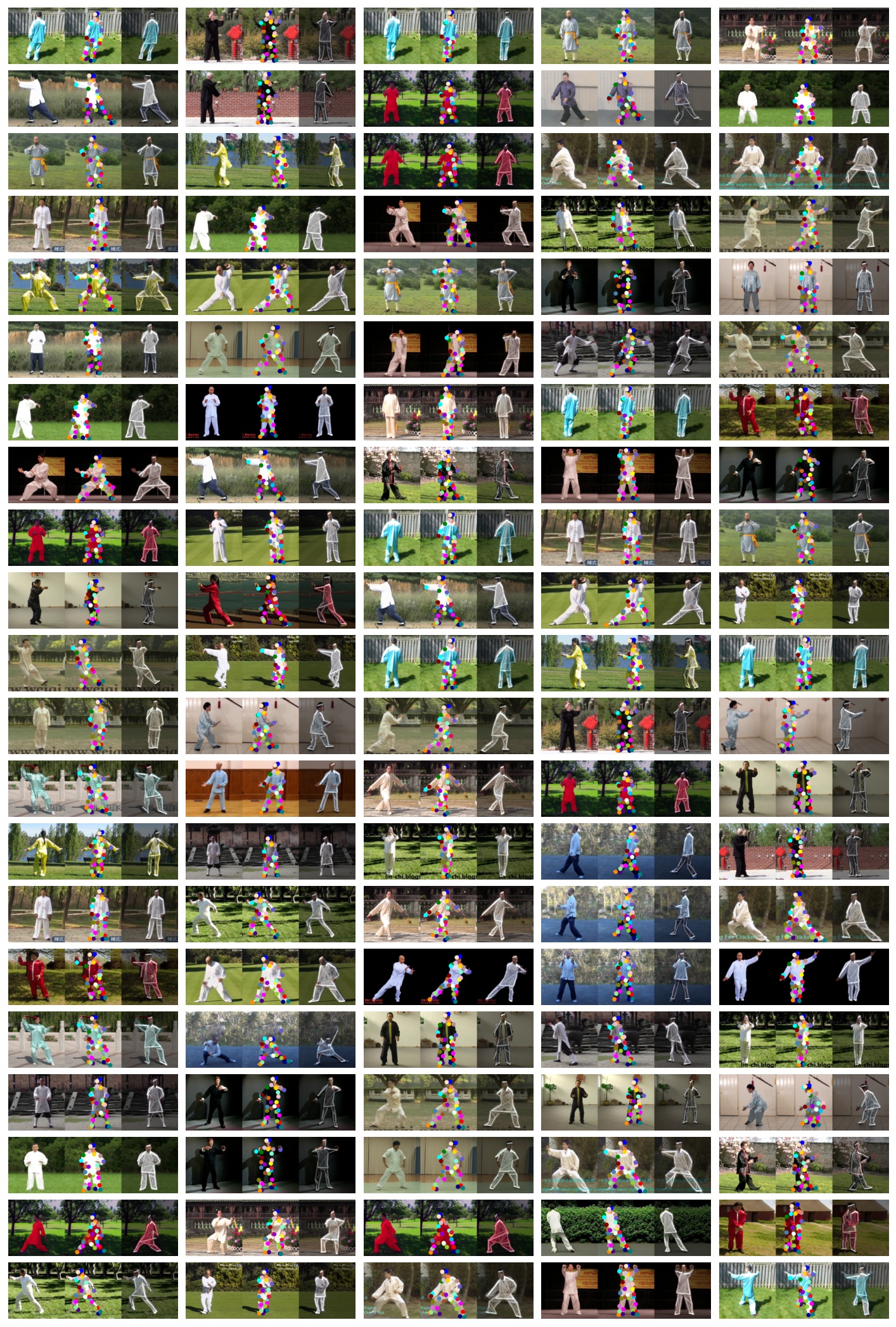}
\end{center}
   \caption{\textbf{105 samples from Taichi (32 keypoints),} with the image-points-edge pairs overlaid.}
   \label{fig:taichi_k32}
\end{figure*}

\end{document}